\documentclass[10pt,journal,comsoc]{IEEEtran}

\usepackage[T1]{fontenc}

\usepackage{cite}

\ifCLASSINFOpdf
\usepackage[pdftex]{graphicx}
\else
\usepackage[dvips]{graphicx}
\fi

\ifCLASSOPTIONcompsoc
\usepackage[caption=false,font=normalsize,labelfon
t=sf,textfont=sf]{subfig}
\else
\usepackage[caption=false,font=footnotesize]{subfi
g}
\fi

\usepackage{amsmath}
\usepackage{mathtools}
\usepackage{amssymb}
\usepackage{mathrsfs}
\usepackage{amsthm}

\DeclarePairedDelimiter{\ceil}{\lceil}{\rceil}

\theoremstyle{plain}
\newtheorem{definition}{Definition}
\newtheorem{theorem}{Theorem}

\newtheoremstyle{cited}%
  {3pt}
  {3pt}
  {\itshape}
  {}
  {\bfseries}
  {.}
  {.5em}
  {\thmname{#1} \thmnumber{#2} \thmnote{\normalfont#3}}

\theoremstyle{cited}
\newtheorem{citedtheorem}[theorem]{Theorem}
\newtheorem{citedlemma}{Lemma}

\usepackage[cmintegrals]{newtxmath}

\usepackage{bm}
\usepackage{bbm}
\usepackage{braket}

\usepackage{algorithm}
\usepackage{algorithmicx}
\usepackage{algpseudocode}

\usepackage{array}
\usepackage{booktabs}

\usepackage{dblfloatfix}

\usepackage{url}
\usepackage{flushend}
\usepackage{color}

\newcommand{\red}[1]{\textcolor{black}{#1}}

\newcommand{\black}[1]{\textcolor{black}{#1}}

\begin{document}
\bstctlcite{IEEEexample:BSTcontrol}

\title{Quantum-Inspired Support Vector Machine}

\author{Chen~Ding,~Tian-Yi~Bao,~and~He-Liang~Huang$^*$
\thanks{This work was supported by the Open Research Fund from State Key Laboratory of High Performance Computing of China (Grant No. 201901-01), National Natural Science Foundation of China under Grants No. 11905294, and China Postdoctoral Science Foundation. (\textit\textbf{Corresponding author: He-Liang Huang. Email: quanhhl@ustc.edu.cn})}

\thanks{Chen Ding is with CAS Centre for Excellence and Synergetic Innovation Centre in Quantum Information and Quantum Physics, University of Science and Technology of China, Hefei, Anhui 230026, China.}
\thanks{Tian-Yi Bao is with Department of Computer Science, University of Oxford, Wolfson Building, Parks Road, OXFORD, OX1 3QD, UK.}
\thanks{He-Liang Huang is with Hefei National Laboratory for Physical Sciences at Microscale and Department of Modern Physics, University of Science and Technology of China, Hefei, Anhui 230026, China, and also with CAS Centre for Excellence and Synergetic Innovation Centre in Quantum Information and Quantum Physics, University of Science and Technology of China, Hefei, Anhui 230026, China.}
}

\markboth{Journal of \LaTeX\ Class Files,~Vol.~14, No.~8, August~2020}{Shell \MakeLowercase{\textit{et al.}}: Bare Demo of IEEEtran.cls for IEEE Communications Society Journals}

\maketitle

\begin{abstract}
Support vector machine (SVM) is a particularly powerful and flexible supervised learning model that analyzes data for both classification and regression, whose usual algorithm complexity scales polynomially with the dimension of data space and the number of data points. \black{To tackle the big data challenge, a quantum SVM algorithm was proposed, which is claimed to achieve exponential speedup  for least squares SVM (LS-SVM). Here, inspired by the quantum SVM algorithm, we present a quantum-inspired classical algorithm for LS-SVM. In our approach, \red{an} improved  fast sampling technique, namely indirect sampling, is proposed for sampling the kernel matrix  and classifying. We first consider the LS-SVM with a linear kernel, and then discuss the generalization of our method to non-linear kernels.} Theoretical analysis shows our algorithm can make classification with arbitrary success probability in logarithmic runtime of both the dimension of data space and the number of data points for low rank, low condition number and high dimensional data matrix, matching the runtime of the quantum SVM.
\end{abstract}


\IEEEpeerreviewmaketitle

\section{Introduction}

\IEEEPARstart{S}{ince} the 1980s, quantum computing has attracted wide attention due to its enormous advantages in solving hard computational problems \cite{huang2020superconducting}, such as integer factorization \cite{shor1994algorithms,lu2007demonstration,huang2017experimental}, database searching \cite{grover1996fast,li2018complementary}, machine learning \cite{qml,huang2018demonstration, liu2019hybrid,huang2017homomorphic, huang2020experimental} and so on \cite{huang2018demonstration2, huang2021emulating}. In 1997, Daniel R. Simon offered compelling evidence that the quantum model may have significantly more complexity theoretic power than the probabilistic Turing machine \cite{simon1997power}. However, it remains an interesting question where is the border between classical computing and quantum computing. Although many proposed quantum algorithms have exponential speedups over the existing classical algorithms, is there any way we can accelerate such classical algorithms to the same complexity of the quantum ones?

In 2018, inspired by the quantum recommendation system algorithm proposed by Iordanis Kerenidis and Anupam Prakash \cite{qRecommendation}, Ewin Tang designed a classical algorithm to produce a recommendation algorithm that can achieve an exponential improvement on previous algorithms \cite{qi_Recommendation}, which is a breakthrough that shows how to apply the subsampling strategy based on Alan Frieze, Ravi Kannan, and Santosh Vempala's 2004 algorithm \cite{FKV} to find a low-rank approximation of a matrix. Subsequently, Tang continued to use same techniques to dequantize two quantum machine learning algorithms, quantum principal component analysis \cite{qpca} and quantum supervised clustering \cite{qsc}, and shows classical algorithms could also match the bounds and runtime of the corresponding quantum algorithms, with only polynomial slowdown \cite{qi_PCA}.

Later, Andr{\'{a}}s Gily{\'{e}}n \textit{et al.}  \cite{regression} and Nai-Hui Chia \textit{et al.} \cite{chia} independently and simultaneously proposed a quantum-inspired matrix inverse algorithm with logarithmic complexity of matrix size, which eliminates the speedup advantage of the famous Harrow-Hassidim-Lloyd (HHL) algorithm \cite{hhl} on certain conditions. Recently, Juan Miguel Arrazola \textit{et al.} studied the actual performance of quantum-inspired algorithms and found that quantum-inspired algorithms can perform well in practice under given conditions. However, the conditions should be further reduced if we want to apply the algorithms to practical datasets \cite{qip}. All of these works give a very promising future
for designing the quantum-inspired algorithm in the machine learning area, where matrix inverse algorithms are universally used.

Support vector machine (SVM) is a data classification algorithm which is commonly used in machine learning area \cite{face,lssvm}. Extensive studies have been conducted on SVMs to boost and optimize their performance, such as the sequential minimal optimization algorithm \cite{platt_sequential_1998}, the cascade SVM algorithm \cite{graf_parallel_2005}, and the SVM algorithms based on Markov sampling \cite{Markov1,Markov2}. These algorithms offer promising speedup either by changing the way of training a classifier, or by reducing the size of training sets. However, the time complexity of current SVM algorithms are all polynomial of data sizes. In 2014, Patrick
Rebentrost, Masoud Mohseni and Seth Lloyd proposed the quantum SVM algorithm \cite{qsvm}, which can achieve an exponential speedup compared to the classical SVMs. The time complexity of quantum SVM algorithm is polynomial of the logarithm of data sizes. Inspired by the quantum SVM algorithm, Tang's methods\cite{qi_Recommendation} and Andr{\'{a}}s Gily{\'{e}}n \textit{et al.}'s work\cite{regression}, we
propose a quantum-inspired classical SVM algorithm, which also shows exponential speedup compared to previous classical SVM for low rank, low condition number and high dimensional data matrix. \black{Both quantum SVM algorithm\cite{qsvm} and our quantum-inspired SVM algorithm are least squares SVM (LS-SVM), which \red{reduce} the optimization problem to finding the solution of a set of linear equations.}

Our algorithm is a dequantization of the quantum SVM algorithm\cite{qsvm}. In quantum SVM algorithm, the \black{labeled data vectors ($x_j$ for $j=1,...,m$)} are mapped to quantum vectors $\ket{x_j}=1/|x_j|\sum(x_j)_k\ket{k}$ via a quantum random access memory (qRAM) and the kernel matrix is prepared using quantum inner product evaluation\cite{qsc}. Then the solution of SVM is found by solving a linear equation system related to the quadratic programming problem of SVM using the quantum matrix inversion algorithm \cite{hhl}. In our quantum-inspired SVM, the labeled vectors are stored in an arborescent data structure which provides the ability to random sampling within logarithmic time of the vector lengths. By performing sampling on these labeled vectors both by their numbers and lengths to get a much smaller dataset, we then find the approximate singular value decomposition of the kernel matrix. And finally, we solve the optimization problem and perform classification based on the solved parameters.

Our methods, particularly the sampling technique, is based on~\cite{qi_Recommendation,regression}. However, the previous sampling techniques cannot be simply copied to solve the SVM tasks, since we don't have an efficient direct sampling access to the kernel matrix we want to perform matrix inversion on (see Section II-B for a more detailed explanation). Hence we have developed an indirect sampling technique to solve such problem. In the whole process, we need to avoid the direct \black{multiplication} on the vectors or matrices with the same size as the kernel, in case losing the exponential speedup. \black{We first consider the LS-SVM with linear kernels, no regularization terms and no bias of the classification hyperplane, which could be regarded as the prototype for quantum-inspired techniques applied in various SVMs. Then we show that the regularization terms can be easily included in the algorithm in Section \ref{sec:main-alg}. Finally, we discuss the generalization of our method to non-linear kernels in Section \ref{subsec:non-linear-svm} and the general case without the constraint on biases of classification hyperplanes in Section \ref{subsec:lssvm}.}
Theoretical analysis shows that our quantum-inspired SVM can achieve exponential speedup over existing classical algorithms under several conditions. Experiments are carried out to demonstrate the feasibility of our algorithm. The indirect sampling developed in our work opens up the possibility of a wider application of the sampling methods into the field of machine learning.

\section{PRELIMINARY}\label{sec:pre}
\subsection{Notations}
We list some matrix-related notations used in this paper.
\begin{table}[h]
\renewcommand{\arraystretch}{1.2}
\label{notation}
\centering
\caption{The Notations}
\begin{tabular}{cl}
\toprule
Symbol & Meaning                                     \\
\midrule
$A$ & matrix $A$ \\
$y$ & vector $y$ or matrix $y$ with only one column   \\
\black{$A^{+}$} & pseudo inverse of $A$       \\
$A^T$ & transpose of $A$ \\
\black{$A^{+T}$} & \black{transpose of pseudo inverse of $A$} \\
$A_{i,*}$ & $i$-th row of $A$   \\
$A_{*,j}$ & $j$-th column of $A$       \\
$\|A\|$ & 2-operator norm of $A$       \\
$\|A\|_F$ & Frobenius norm of $A$       \\
$Q(\cdot)$ & time complexity for querying an element of $\cdot$   \\
$L(\cdot)$ & time complexity for sampling an element of $\cdot$   \\
\bottomrule
\end{tabular}
\end{table}
\subsection{Least squares SVM}\label{subsection_svm}

Suppose we have $m$ data points $\{(x_j,y_j):x_j\in\mathbb{R}^n, y_j=\pm 1\}_{j=1,\dots,m}$, where $y_j=\pm 1$ depending on the class which $x_j$ belongs to. Denote $(x_1,...,x_m)$ by $X$ and $(y_1,\dots,y_m)^T$ by $y$. A SVM finds a pair of parallel hyperplanes $x\cdot w+b=\pm 1$ that divides the points into two classes depending on the given data. Then for any new input points, it can make classification by its relative position with the hyperplanes.

\black{
We make \red{the} following assumption on the dataset so as to simplify the problem: Assume these data points are equally distributed on both sides of a hyperplane that passes through the origin and their labels are divided by such hyperplane. Thus we assume $b=0$. An generalized method for $b\neq 0$ is discussed in Section~\ref{subsec:lssvm}.
}

According to \cite{lssvm}, the optimization problem of \black{LS-SVM with linear kernel} is
\begin{gather*}
  \min\limits_{w,b,e} \mathcal{L}_1(w,b,e)=\frac{1}{2}w^Tw+\frac{\gamma}{2}\sum^m_{k=1}e^2_k,\\
\text{subject to}\quad y_k(w^Tx_k+b)=1-e_k,\quad k=1,\dots,m.
\end{gather*}

Take $b=0$, we get
\begin{gather*}
  \min\limits_{w,e} \mathcal{L}_2(w,e)=\frac{1}{2}w^Tw+\frac{\gamma}{2}\sum^m_{k=1}e^2_k,\\
\text{subject to}\quad y_kw^Tx_k=1-e_k,\quad k=1,\dots,m.
\end{gather*}

One defines the Lagrangian
\[\mathscr{L}(w,e,\mu)=\mathcal{L}_2(w,e)-\sum^m_{k=1}\mu_k (y_kw^Tx_k-1+e_k).\]

The condition for optimality
\begin{align*}
  \frac{\partial \mathscr{L}}{\partial w}=0\rightarrow&w=\sum^m_{k=1}\mu_ky_kx_k,\\
  \frac{\partial \mathscr{L}}{\partial e_k}=0\rightarrow&\mu_k=\gamma e_k, k=1,\dots,m,\\
  \frac{\partial \mathscr{L}}{\partial \mu_k}=0\rightarrow&y_kw^Tx_k-1+e_k=0, k=1,\dots,m
\end{align*}
can be written as the solution to the following set of linear equations $Z^TZ\mu+\gamma^{-1}\mu=1$, where $Z=(x_1y_1,\dots,x_my_m)$. Let $\alpha_k=\mu_ky_k$, we have
\begin{gather}\label{eqn1}
  (X^TX+\gamma^{-1}I)\alpha=y.
\end{gather}
Once $\alpha$ is solved, the classification hyperplane will be $x^TX\alpha=0$. Given query point $x$, we evaluate $\text{sgn}(x^TX\alpha)$ to make classification.

We use our sampling techniques in solving Equation~(\ref{eqn1}) and evaluating $\text{sgn}(x^TX\alpha)$ to avoid time complexity overhead of $\text{poly}(m)$ or $\text{poly}(n)$, which will kill the wanted exponential speedup. Note that the quantum-inspired algorithm for linear equations \cite{regression,chia} may inverse a low-rank matrix in logarithmic runtime. However, such algorithm cannot be invoked directly to solve Equation~(\ref{eqn1}) here, since the complexity of direct computing the matrix \black{$X^TX+\gamma^{-1}I$} is polynomial, which would once again kill the exponential speedup. Thus we need to develop the indirect sampling technique to efficiently perform matrix inversion on \black{$X^TX+\gamma^{-1}I$} with only sampling access of $X$.

\subsection{The sampling technique}\label{subsection_sample}
We show the definition and idea of our sampling method to get indices, elements or submatrices, which is the key technique used in our algorithm, as well as in \cite{FKV,qi_Recommendation,regression}.
\begin{definition}[Sampling on vectors]
Suppose $v\in \mathbb{C}^n$, define $q^{(v)}$ as a probability distribution that:
\begin{displaymath}
x\sim q^{(v)}: \quad \mathbb{P}[x=i]=\frac{|v_i|^2}{\|v\|^2}.
\end{displaymath}
Picking an index according to the probability distribution $q^{(v)}$ is called a sampling on $v$.
\end{definition}

\begin{definition}[Sampling the indices from matrices]
Suppose $A\in \mathbb{C}^{n\times m}$, define $q^{(A)}$ as a 2-dimensional probability distribution that:
\begin{displaymath}
(x,y)\sim q^{(v)}: \quad \mathbb{P}[x=i,y=j]=\frac{|A_{ij}|^2}{\|A\|_F^2}.
\end{displaymath}
Picking a pair of indices $(i,j)$ according to the probability distribution $q^{(A)}$ is called a sampling on $A$.
\end{definition}

\begin{figure*}[!t]
  \normalsize
  \centering
  \fbox{\includegraphics[width=7in]{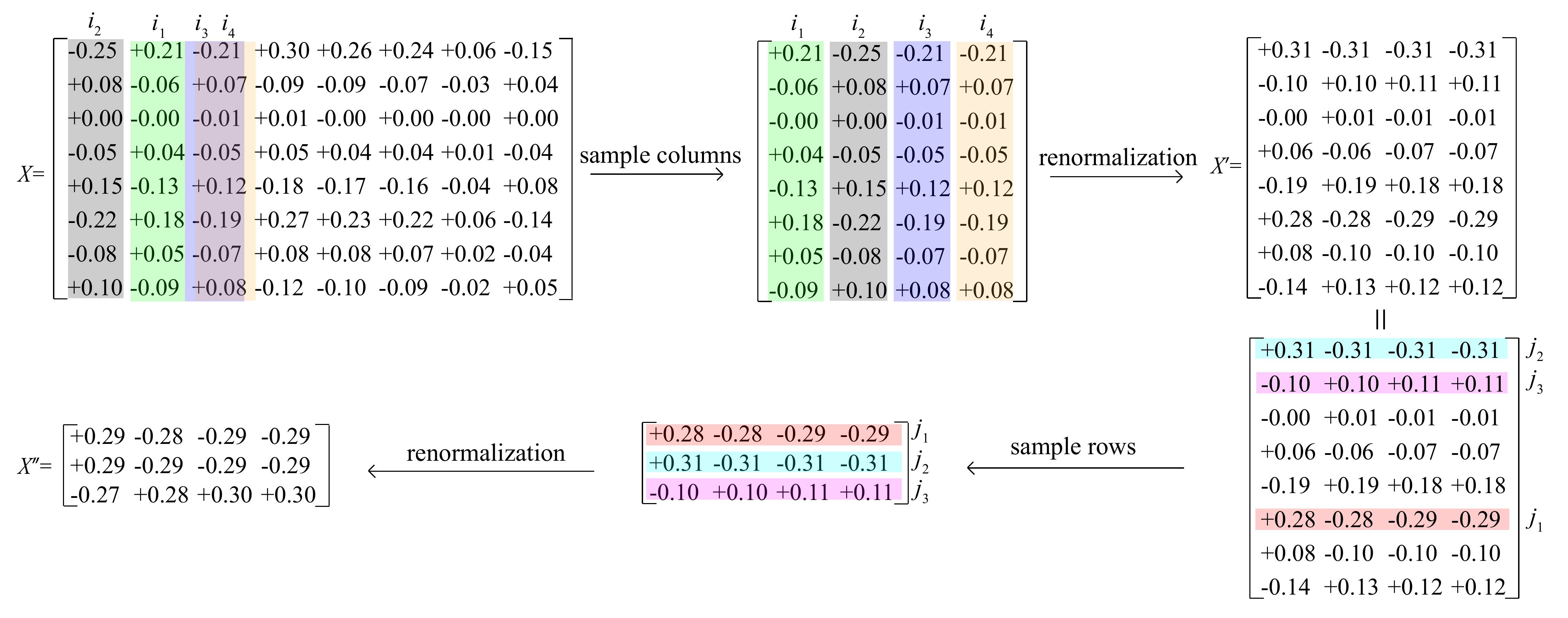}}
  \caption{A demonstration of sampling submatrices from matrices (The process described in Def. 3, which is also Step 2 and Step 3 in Alg. \ref{main_alg}.). We sample columns from $X$ to get $X'$ and sample rows from $X'$ to get $X''$. Note that $X'$ and $X''$ are normalized such that $\mathbb{E}[X'X^{\prime T}]=XX^T$ and $\mathbb{E}[X^{\prime\prime T}X'']=X^{\prime T}X'$.}
  \label{fig:subsampling}
  \hrulefill
  \vspace*{4pt}
  \end{figure*}

\begin{definition}[Sampling the submatrices from matrices]
Suppose the target is to sample a submatrix $X''\in \mathbb{C}^{c\times r}$ from $X\in \mathbb{C}^{n\times m}$. \black{First} we sample $r$ times on the vector $(\|X_{*,j}\|)_{j=1,...,m}$ and get column indices $j_1,...,j_r$. The columns $X_{*,j_1},...,X_{*,j_r}$ form submatrix $X'$. Then we sample $c$ times on the $j$-th column of $X$ and get row indices $i_1,... ,i_c$. In each time the $j$ is \black{sampled uniformly at random} from $j_1,... ,j_r$. The rows $X'_{i_1,*},...,X'_{i_c}$ form submatrix $X''$. The matrices $X'$ and $X''$ are normalized so that $\mathbb{E}[X'X^{\prime T}]=XX^T$ and $\mathbb{E}[X^{\prime\prime T}X'']=X^{\prime T}X'$.
\end{definition}

The process of sampling the submatrices from matrices (as described in Def.~3) is shown in Fig.~\ref{fig:subsampling}. To put it simple, it is taking several rows and columns out of the matrix by a random choice decided by the ``importance'' of the elements. Then \black{normalize} them so that they are unbiased from the original rows and columns.

To achieve fast sampling, we usually store vectors in an arborescent data structure (such as binary search tree) as suggested in \cite{qi_Recommendation} and store matrices by a list of their row trees or column trees. Actually, the sampling is an analog of quantum states measurements. It only reveals a low-dimensional projection of vectors and matrices in each calculation. Rather than computing with the whole vector or matrix, we choose the most representative elements of them for calculation with a high probability (we choose the elements according to the probability of their squares, which is also similar to the quantum measurement of quantum states.). The sampling technique we use has the advantage of unbiasedly representing the original vector while consuming less computing resources.

We note that there are other kinds of sampling methods for SVM such as the Markov sampling \cite{Markov1,Markov2}. Different sampling methods may work well on different scenarios. Our algorithm is designed for low-rank datasets, while the algorithms based on Markov sampling \cite{Markov1,Markov2} may work well on the datasets that the columns form a uniformly ergodic Markov chain. In our algorithm, to achieve exponential speedup, the sampling technique is different from Markov sampling: (i) We sample both the rows and columns of matrix, rather than only sampling columns. (ii) We sample each elements according to norm-squared probability distribution. (iii) In each dot product calculation (Alg.~\ref{estimation_alg}), we use sampling technique to avoid operations with high complexity.

\subsection{The preliminary algorithms}
We invoke two algorithms employing sampling techniques for saving complexity from \cite{regression}. They are treated as oracles that outputs certain outcomes with controlled errors in the main algorithm. \black{Lemma \ref{lemma1} and Lemma \ref{lemma2} shows their correctness and efficiency. For the sake of convenience, some minor changes on the algorithms and lemmas are  made.}

\subsubsection{Trace inner product estimation}
Alg. \ref{estimation_alg} achieves calculation of trace inner products with logarithmic time on the sizes of the matrices.

\begin{algorithm}[H]
\centering
\begin{algorithmic}[1]
\item[\textbf{Input:}] $A\in \mathbb{C}^{m\times n}$ that we have sampling access in complexity $L(A)$ and $B\in \mathbb{C}^{n\times m}$ that we have query access in complexity $Q(B)$. Relative error bound $\xi$ and success probability bound $1-\eta$.
\item[\textbf{Goal:}] Estimate $\text{Tr}[AB]$.
\State Repeat step \ref{alg2:step2} $\ceil{6\log_2(\frac{2}{\eta})}$ times and take the median of $Y$, noted as $Z$.
\State Repeat step \ref{alg2:step1} $\ceil{\frac{9}{\xi^2}}$ times and calculate the mean of $X$, noted as $Y$.\label{alg2:step2}
\State Sample $i$ from row norms of $A$. Sample $j$ from $A_i$. Let $X=\frac{\|A\|^2_F}{A_{ij}}B_{ji}$.\label{alg2:step1}
\item[\textbf{Output:}] $Z$.
\end{algorithmic}
\caption{Trace Inner Product Estimation.}
\label{estimation_alg}
\end{algorithm}

\black{
\begin{citedlemma}[\cite{regression}]\label{lemma1}
  Suppose that we have length-square sampling access to $A\in\mathbb{C}^{m\times n}$ and query access to the matrix $B\in\mathbb{C}^{n\times m}$ in complexity $Q(B)$. Then we can estimate {\rm \text{Tr}}$[AB]$ to precision $\xi\|A\|_F\|B\|_F$ with probability at least $1-\eta$ in time
  \[O\left(\frac{\log(1/\eta)}{\xi^2}(L(A)+Q(B))\right).\]
\end{citedlemma}
}
\begin{algorithm}[H]
  \centering
  \begin{algorithmic}[1]
  \item[\textbf{Input:}] $A\in \mathbb{C}^{m\times n}$ that we have length-square sampling access and $b\in \mathbb{C}^{n}$ that we have norm access and $y=Ab$ that we have query access.
  \item[\textbf{Goal:}] Sample from length-square distribution of $y=Ab$.
  \State Take $D\geq\|b\|^2$.
  \State Sample a row index $i$ by row norm squares of $A$.\label{alg3:last}
  \State Query $|y_i|^2=|A_{i,*}b|^2$ and calculate $\frac{|A_{i,*}b|^2}{D\|A_{i,*}\|^2}$.
  \State Sample a real number $x$ uniformly distributed in $[0,1]$. If $x<\frac{|A_{i,*}b|^2}{D\|A_{i,*}\|^2}$, output $i$, else, go to step \ref{alg3:last}.
  \item[\textbf{Output:}] The row index $i$.
  \end{algorithmic}
  \caption{Rejection sampling.}
  \label{rejection_alg}
  \end{algorithm}
\subsubsection{Rejection sampling}
Alg. \ref{rejection_alg} achieves sampling of a vector that we do not have full query access in time logarithmic of its length.

\black{
\begin{citedlemma}[\cite{regression}]\label{lemma2}
  Suppose that we have length-square sampling access to $A\in\mathbb{C}^{m\times n}$ having normalized rows, and we are given $b\in\mathbb{C}^n$. Then we can implement queries to the vector $y:=Ab\in \mathbb{C}^n$ with complexity $Q(y)=O(nQ(A))$ and we can length-square sample from $q^{(y)}$ with complexity $L(y)$ such that $\mathbb{E}[L(y)]=O\left(\frac{n\|b\|^2}{\|y\|^2}(L(A)+nQ(A))\right)$.
\end{citedlemma}
}

\section{Quantum-inspired SVM Algorithm}\label{sec:main-alg}
\begin{figure*}[!t]
  \normalsize
  \centering
  \fbox{\includegraphics{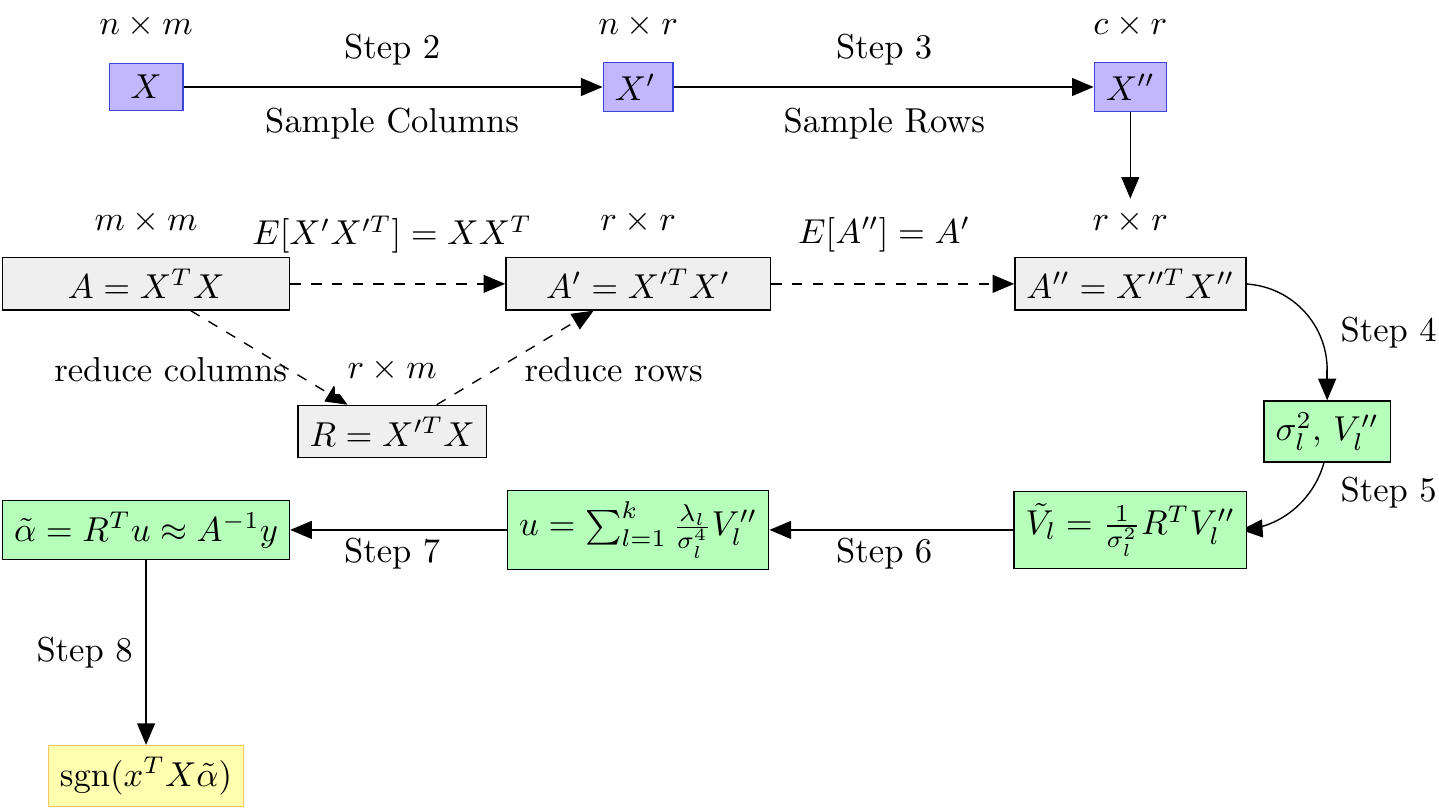}}
  \caption{The quantum-inspired SVM algorithm. In the algorithm, the subsampling of $A$ is implemented by subsampling the matrix $X$ (Step 1-3), which is called the indirect sampling technique. After the indirect sampling, we perform the spectral decomposition (Step 4). Then we estimate the approximation of the eigenvectors ($\tilde{V}_l$) of $A$ (Step 5). Finally, we estimate the classification expression (Step 6-8).}
  \label{fig:flow}
  \hrulefill
  \vspace*{4pt}
  \end{figure*}

We show the main algorithm (Alg. \ref{main_alg}) that makes classification as the classical SVMs do. Note that actual calculation only happens when we use the expression "calculate" in this algorithm. Otherwise it will lose the exponential-speedup advantage for operations on large vectors or matrices. $\gamma$ is temporarily taken as $\infty$. Fig.~\ref{fig:flow} shows the algorithm process.
\begin{algorithm}[H]
\centering
\caption{Quantum-inspired SVM Algorithm.}\label{main_alg}
\begin{algorithmic}[1]
\item[\textbf{Input:}] $m$ training data points and their labels $\{(x_j,y_j):x_j\in\mathbb{R}^n, y_j=\pm 1\}_{j=1,\dots,m}$, where $y_j=\pm 1$ depending on the class to which $x_j$ belongs. Error bound $\epsilon$ and success probability bound $1-\eta$. $\gamma$ set as $\infty$.
\item[\textbf{Goal 1:}] Find $\tilde{\alpha}$ that $\|\tilde{\alpha}-\alpha\|\leq \epsilon\|\alpha\|$ with success probability at least $1-\eta$, in which $\alpha=(X^TX)^{+}y$.
\item[\textbf{Goal 2:}] For any given $x\in \mathbb{R}^n$, find its class.
\State\textbf{Init:}\label{step:init} Set $r,c$ as described in (\ref{eq:r}) and (\ref{eq:c}).
\State\textbf{Sample columns:}\label{step:sample_cols} Sample $r$ column indices $i_1, i_2,...,i_r$ according to the column norm squares $\frac{\|X_{*,i}\|^2}{\|X\|^2_F}$. Define $X'$ to be the matrix whose $s$-th column is $\frac{\|X\|_F}{\sqrt{r}}\frac{X_{*,i_s}}{\|X_{*,i_s}\|}$. Define $A'=X^{\prime T}X'$.
\algstore{bkbreak}
\end{algorithmic}
\end{algorithm}
\begin{algorithm}[H]
  \addtocounter{algorithm}{-1}
\centering
\caption{Quantum-inspired SVM Algorithm.}\label{main_alg2}
\begin{algorithmic}[1]
\algrestore{bkbreak}
\State\textbf{Sample rows:}\label{step:sample_rows} Sample $s\in [r]$ uniformly, then sample a row index $j$ distributed as $\frac{|X'_{js}|^2}{\|X'_{*,s}\|^2}$. Sample a total number of $c$ row indices $j_1,j_2,...,j_c$ this way. Define $X''$ whose $t$-th row is $\frac{\|X\|_F}{\sqrt{c}}\frac{X'_{j_t,*}}{\|X'_{j_t,*}\|}$. Define $A''=X^{\prime\prime T}X''$.

\State\textbf{Spectral decomposition:}\label{main_alg:sd} Calculate the spectral decomposition of $A''$. Denote here by $A''=V^{\prime\prime}\Sigma^2V^{\prime\prime T}$. Denote the calculated eigenvalues by $\sigma_l^2$, $l=1,\dots,k$.

\State\textbf{Approximate eigenvectors:}\label{step:appeigen} Let $R=X^{\prime T}X$. Define $\tilde{V}_l=\frac{R^TV''_l}{\sigma_l^2}$ for $l=1,...,k$, $\tilde{V}=(\tilde{V}_l)_{l=1,...,k}$.

\State\textbf{Estimate matrix elements:}\label{step:elements} Calculate $\tilde{\lambda_l}=\tilde{V}_l^Ty$ to precision $\frac{3\epsilon\sigma_l^2}{16\sqrt{k}}\|y\|$ by Alg. \ref{estimation_alg}, each with success probability $1-\frac{\eta}{4k}$. Let $u=\sum^k_{l=1}\frac{\tilde{\lambda_l}}{\sigma_l^4}V''_l$.

\State\textbf{Find query access:}\label{step:find_query} Find query access of $\tilde{\alpha}=\tilde{R}^Tu$ by $\tilde{\alpha}_p=u^T\tilde{R}_{*,p}$, in which $\tilde{R}_{ij}$ is calculated to precision $\frac{\epsilon\kappa^2}{4\|X\|_F}$ by Alg.~\ref{estimation_alg}, each with success probability $1-\frac{\eta}{4\ceil{864/\epsilon^2\log(8/\eta)}}$.

\State\textbf{Find sign:}\label{step:find_sign} Calculate $x^TX\tilde{\alpha}$ to precision $\frac{\epsilon}{4}\|\alpha\|\|x\|$ with success probability $1-\frac{\eta}{4}$ by Alg. \ref{estimation_alg}. Tell its sign.
\item[\textbf{Output:}] The answer class depends on the sign. \black{Positive} corresponds to 1 while negative to $-1$.
\end{algorithmic}
\end{algorithm}

The following theorem states the accuracy and time complexity of quantum-inspired support vector machine algorithm, from which we conclude the time complexity $T$ depends polylogarithmically on $m,n$ and polynomially on $k,\kappa,\epsilon,\eta$. It is to be proved in section \ref{sec:accuracy} and section \ref{sec:complexity}.

\begin{theorem}
Given parameters \black{$\epsilon>0,0<\eta<1$,} and given the data matrix $X$ with size $m \times n$, rank $k$, norm $1$, and condition number $\kappa$, the quantum-inspired SVM algorithm will find the classification expression $x^TX\alpha$ for any vector $x\in \mathbb{C}^{n}$ with error less than $\epsilon\kappa^2\sqrt{m}\|x\|$, success probability higher than $1-\eta$ and time complexity $T(m,n,k,\kappa,\epsilon,\eta)$.
\begin{align*}
T&=O(r\log_2m+cr\log_2n+r^3\\
&+\frac{\|X\|_F^2k^2}{\epsilon^2}\log_2(\frac{8k}{\eta})(\log_2(mn)+k)\\
&+\frac{1}{\epsilon^2}\log_2\frac{1}{\eta}(\log_2(mn)+rk\log_2(\frac{2}{\eta_1})\frac{\|X\|^4_F}{\epsilon_1^2r}\log_2(mn))),
\end{align*}
in which
\begin{displaymath}
  \epsilon_1=\frac{\epsilon\|x\|}{2\sqrt{r}\ceil{\frac{36}{\epsilon^2}}\ceil{6\log_2(\frac{16}{\eta})}},
\end{displaymath}
  \begin{displaymath}
\eta_1=\frac{\eta}{8r\ceil{\frac{36}{\epsilon^2}}\ceil{6\log_2(\frac{16}{\eta})}}.
\end{displaymath}
\end{theorem}

\black{
In Alg.~\ref{main_alg}, $\gamma$ is set as $\infty$, which makes the coefficient matrix $A=X^TX$. Notice that the eigenvectors of $X^TX+\gamma^{-1}I$ and $X^TX$ are the same, and the difference of their eigenvalues are $\gamma^{-1}$. Thus the algorithm can be easily extended to be applied to the coefficient matrix $X^TX+\gamma^{-1}I$ with arbitrary $\gamma$, by just simply adding $\gamma^{-1}$ to the calculated eigenvalues in Step~\ref{main_alg:sd}.}

\section{Accuracy}\label{sec:accuracy}
We prove that the error of computing the classification expression $x^TX\tilde{\alpha}$ in the quantum-inspired SVM algorithm will not exceed $\epsilon\kappa^2\sqrt{m}\|x\|$.
We take $\gamma=\infty$ in the analysis because adding $\gamma^{-1}$ to the eigenvalues won't cause error and thus the analysis is the same in the case of $\gamma\neq\infty$. \red{We first show how to break the total error into multiple parts, and then analyze each part in the subsections.}

Let $\alpha=(X^TX)^{+}y$, $\alpha'=\sum^k_{l=1}\frac{\lambda_l}{\sigma_l^2}\tilde{V}_l=\tilde{V}\Sigma^{-2}\tilde{V}^Ty$, in which $\lambda_l=\tilde{V}^T_ly$ and $\alpha''=\sum^k_{l=1}\frac{\tilde{\lambda}_l}{\sigma_l^2}\tilde{V}_l$.
Then the total error of the classification expression is \footnote{\black{For any expression $f$, $\Delta(f)$ represents the difference of the exact value of $f$ and the value calculated by the estimation algorithms Alg.~\ref{estimation_alg} and Alg.~\ref{main_alg} (These two algorithms cannot get the exact values because randomness is introduced.)}.}

\begin{align*}
E&= \Delta(x^TX\alpha) \\
    &\leq |x^TX(\alpha-\tilde{\alpha})|+\Delta(x^TX\tilde{\alpha})\\
    &\leq \|x\|(\|\alpha-\alpha'\|+\|\alpha'-\alpha''\|+\|\alpha''-\tilde{\alpha}\|)+\Delta(x^TX\tilde{\alpha})
\end{align*}

\red{Denote $E_1=\|x\|\|\alpha'-\alpha\|$, $E_2=\|x\|\|\alpha''-\alpha'\|$, $E_3=\|x\|\|\tilde{\alpha}-\alpha''\|$, $E_4=\Delta(x^TX\tilde{\alpha})$. Our target is to show each of them is no more than $\frac{\epsilon}{4} \|\alpha\|\|x\|$  with probability no less than $1-\frac{\eta}{4}$. So that
\begin{align*}
  E &\leq E_1+E_2+E_3+E_4\\
  &\leq \epsilon\kappa^2\sqrt{m} \|x\|,
  \end{align*}
  with success probability no less than $1-\eta$.
}

\red{$E_1$ represents the error introduced by subsampling and eigenvector approximation (i.e., Step 1-5 in Alg. \ref{main_alg}).} The fact that it is less than $\frac{\epsilon}{4} \|\alpha\|\|x\|$ with probability no less than $1-\frac{\eta}{4}$ is shown in subsection \ref{sec:err2}.

\red{$E_2$ represents the error introduced by approximation on $\lambda_l$ (i.e., Step 6 in Alg. \ref{main_alg}).} The fact that it is less than $\frac{\epsilon}{4} \|\alpha\|\|x\|$ with probability no less than $1-\frac{\eta}{4}$ is shown in subsection \ref{sec:err1}.

\red{$E_3$ represents the error introduced in query of $R$ and $\alpha$.} The fact that it is less than $\frac{\epsilon}{4} \|\alpha\|\|x\|$ with probability no less than $1-\frac{\eta}{4}$ is guaranteed by Step \ref{step:find_query} of Alg. \ref{main_alg}.

\red{$E_4$ represents the error caused by Alg. \ref{estimation_alg} in estimating $x^TX\tilde{\alpha}$ as the footnote${^1}$ suggests.} The fact that it is less than $\frac{\epsilon}{4} \|\alpha\|\|x\|$ with probability no less than $1-\frac{\eta}{4}$ is guaranteed by Step \ref{step:find_sign} of Alg. \ref{main_alg}.

For achieving accurate classification, we only need a relative error $\frac{E}{x^TX\alpha}$ less than $1$. Thus \black{by lessening} $\epsilon$, we can achieve this goal in any given probability range.
\subsection{Proof of $E_2\leq \frac{\epsilon}{4} \|\alpha\|\|x\|$}\label{sec:err1}
Notice that
\begin{align*}
E_3 &= \|x\|\|\alpha-\alpha'\| \\
&= \|x\|\|\alpha-\tilde{V}\Sigma^{-2}\tilde{V}^TA\alpha\| \\
&\leq \|\alpha\|\|x\|\|\tilde{V}\Sigma^{-2}\tilde{V}^TA-I_m\|.
\end{align*}

\begin{figure}[!t]
  \normalsize
  \centering
  \fbox{\includegraphics{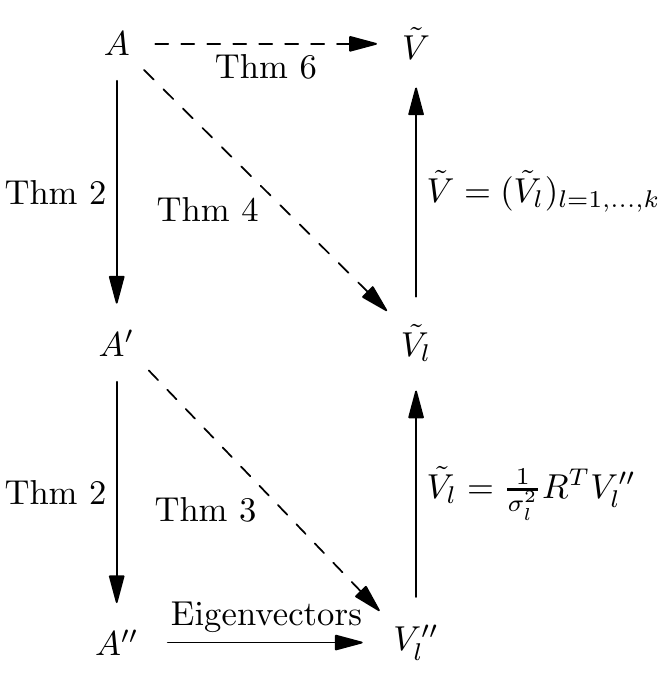}}
  \caption{The whole procedure of proving $\|\tilde{V}\Sigma^{-2}\tilde{V}^TA-I_m\|\leq\frac{\epsilon}{2}$. Thm 2 shows the difference among $A$ and the subsampling outcomes $A'$ and $A''$. Thm 3 shows the relation between $A'$ and $V''_l$. Thm 4 shows the relation between $A$ and $\tilde{V}_l$. Thm 6 shows the final relation between $A$ and $\tilde{V}$.}
  \label{fig2}
  \hrulefill
  \vspace*{4pt}
  \end{figure}

Here we put 5 theorems (from \ref{theorem:unbiased_matrix} to \ref{thm:final_err}) to prove $\|\tilde{V}\Sigma^{-2}\tilde{V}^TA-I_m\|\leq\frac{\epsilon}{4}$, in which theorem \ref{theorem:unbiased_matrix} and \ref{thm:err_of_B} are invoked from \cite{regression}. We offer proofs for Theorem \ref{thm1},\ref{thm:to_prove} and \ref{thm:final_err} in appendix \ref{app:proof}. The purpose of these theorems is to show that $\tilde{V}\Sigma^{-2}\tilde{V}^T $ is functionally close to the inverse of matrix A, as $\|\tilde{V}\Sigma^{-2}\tilde{V}^TA-I_m\|\leq\frac{\epsilon}{4}$ suggests.

Theorem \ref{theorem:unbiased_matrix} states the norm distance between $A$, $A'$ and $A''$. According to the norm distance, and the fact that $V''_l$ are the eigenvectors of $A''$, Theorem \ref{thm1} finds the relation between $A'$ and $V''_l$. We define $\tilde{V}_l=\frac{1}{\sigma^2_l}R^TV''_l$, and Theorem \ref{thm:final_err} finally gives the relation between $A$ and $\tilde{V}$. The procedure is shown in Fig.~\ref{fig2}.

\begin{citedtheorem}[\black{\cite{regression}}]\label{theorem:unbiased_matrix}
Let $X'\in \mathbb{C}^{n\times r}$, $X''\in \mathbb{C}^{c\times r}$ is the sampling outcome of $X'$. Suppose $X''$ is normalized that $\mathbb{E}[X^{\prime\prime T}X'']=X^{\prime T}X'$, then $\forall \epsilon \in [0,\frac{\|X'\|}{\|X'\|_F}]$, we have
\begin{displaymath}
\mathbb{P}\left[\|X^{\prime T}X'-X^{\prime \prime T}X''\|\geq \epsilon\|X'\|\|X'\|_F\right]\leq 2re^{-\frac{\epsilon^2c}{4}}.
\end{displaymath}

Hence, for $c\geq \frac{4\log_2 (\frac{2r}{\eta})}{\epsilon^2}$, with probability at least $1-\eta$ we have
\begin{displaymath}
\|X^{\prime T}X'-X^{\prime \prime T}X''\|\leq \epsilon\|X'\|\|X'\|_F.
\end{displaymath}
\end{citedtheorem}

When a submatrix $X''$ is randomly subsampled from $X'$, it is a matrix of multiple random variables. Theorem \ref{theorem:unbiased_matrix} is the Chebyshev's Inequality for $X''$. It points out that the operator norm distance between $X^{\prime T}X'$ and $X^{\prime\prime T}X''$ is short with a high probability.

\begin{theorem}\label{thm1}
Suppose \red{the columns of matrix $V''$, denoted as $V''_l, l=1,\dots,k$, are orthogonal normalized vectors} while \begin{displaymath}A''=\sum^k_{l=1}\sigma_l^2V''_lV^{\prime\prime T}_l.\end{displaymath}
Suppose $\|A'-A''\|\leq\beta$. Then \red{$\forall i,j\in\{1,...,r\}$},
\begin{displaymath}
|V_i^{\prime \prime T}A'V''_j-\delta_{ij}\sigma_i^2|\leq\beta.
\end{displaymath}
\end{theorem}

Theorem \ref{thm1} points out that if matrix $A'$ and $A''$ are close in operator norm sense, $A''$'s eigenvectors will approximately work as eigenvectors for $A'$ too.

\begin{theorem}\label{thm:to_prove}
Suppose \red{the columns of matrix $V''$, denoted as $V''_l, l=1,\dots,k$, are orthogonal normalized vectors} while
\begin{displaymath}
|V_i^{\prime \prime T}A'V''_j-\delta_{ij}\sigma_i^2|\leq\beta,\quad \red{\forall i,j\in\{1,...,r\}}.
\end{displaymath}
Suppose $\|XX^T-X'X^{\prime T}\|
\leq \epsilon' $, $\|X\|\leq 1$, $\frac{1}{\kappa}\leq \sigma_i^2\leq 1$ and the condition of Thm \ref{thm1} suffices. Let $\tilde{V}_l=\frac{R^TV''_l}{\sigma_l^2}$, then
\begin{displaymath}
|\tilde{V}_i^T\tilde{V}_j-\delta_{ij}|\leq\kappa^2\beta^2+2\kappa\beta+\kappa^2\epsilon'\|X\|^2_F,
\end{displaymath}
and
\begin{align*}
|\tilde{V}_i^TA\tilde{V}_j-\delta_{ij}\sigma^2_i|\leq(2\epsilon'+\beta\|X\|^2_F)\|X\|^2_F\kappa^2.
\end{align*}
in which $A'=X^{\prime T}X'$, $A=X^TX$.
\end{theorem}

Theorem \ref{thm:to_prove} points out that if $A''$'s eigenvectors approximately work as eigenvectors for $A'$ and $\|XX^T-X'X^{\prime T}\|
\leq \epsilon' $, $\tilde{V}^T_l$ approximately work as eigenvectors for $A$.

\begin{citedtheorem}[\black{\cite{regression}}]\label{thm:err_of_B}
If $\text{rank}(B)\leq k$, $\tilde{V}$ has $k$ columns that spans the row and column space of $B$, then
\begin{displaymath}\|B\|\leq \|(\tilde{V}^T\tilde{V})^{+}\|\|\tilde{V}^TB\tilde{V}\|.\end{displaymath}
\end{citedtheorem}

Under the condition that $\tilde{V}^T_l$ approximately work as eigenvectors for $A$, the following Theorem \ref{thm:final_err} points out that $\tilde{V}\Sigma^{-2}\tilde{V}^T $ is functionally close to the inverse of matrix A.

\black{
\begin{theorem}\label{thm:final_err}
If \red{$\forall i,j\in\{1,\dots,k\},$}
\begin{equation}\label{eq:err_of_approximated_orthogonals}
|\tilde{V}_i^T\tilde{V}_j-\delta_{ij}|\leq \frac{1}{4k},
\end{equation}
\begin{displaymath}
|\tilde{V}_i^TA\tilde{V}_j-\delta_{ij}\sigma_i^2|\leq \zeta,
\end{displaymath}
and the condition of Thm \ref{thm:to_prove} suffices.
Then
\begin{displaymath}\|\tilde{V}\Sigma^{-2}\tilde{V}^TA-I_m\|\leq \frac{5}{3}\kappa k\zeta.\end{displaymath}
\end{theorem}
}

To conclude, for $\mathbb{P}[\|\alpha'-\alpha\|>\frac{\epsilon}{4}\|\alpha\|]\leq \frac{\eta}{4}$, we need to pick $\epsilon'$ and $\beta$ such that
\begin{align}\label{eq3}
  \kappa^2\beta^2+2\kappa\beta+\kappa^2\epsilon'\|X\|^2_F\leq\frac{1}{4k},
\end{align}
\begin{align}\label{eq4}
  (2\epsilon'+\beta\|X\|^2_F)\|X\|^2_F\kappa^2\leq \zeta,
\end{align}
\begin{align}\label{eq5}
\frac{5}{3}\kappa k\zeta\leq\frac{\epsilon}{4},
\end{align}
and decide the sampling parameter as
\begin{align}
r&=\ceil{\frac{4\log_2(\frac{8n}{\eta})}{\epsilon'^2}},\label{eq:r}
\\
c&=\ceil{ \frac{4\kappa^2\log_2(\frac{8r}{\eta})}{\beta^2}}\label{eq:c}.
\end{align}
\subsection{Proof of $E_1\leq \frac{\epsilon}{4} \|\alpha\|\|x\|$}\label{sec:err2}
Notice that
\begin{align*}
E_4 &= \|x\|\|\alpha-\tilde{\alpha}\|.
\end{align*}
For $y=X^TX\alpha$ and $\alpha=X^{+}X^{+T}y$, we have $\|y\|\leq \|\alpha\|\leq \kappa^2\|y\|$.

For $\|\tilde{\alpha}-\alpha'\|$, let $z$ be the vector that $z_l=\frac{\lambda_l-\tilde{\lambda_l}}{\sigma_l^2}$, we have
\begin{align*}
\|\tilde{\alpha}-\alpha'\| =&\|\sum^k_{l=1}\frac{\lambda_l-\tilde{\lambda_l}}{\sigma_l^2}\tilde{V}_l\|\\
=&\|\tilde{V}z\|\\
\leq& \sqrt{\|\tilde{V}^T\tilde{V}\|}\|z\|\\
\leq& \frac{4}{3}\frac{3\epsilon\sigma_l^2}{8\sqrt{k}}\|y\|\frac{1}{\sigma_l^2}\sqrt{k}\\
\leq& \frac{1}{4}\epsilon\|\alpha\|.
\end{align*}
in which $\|\tilde{V}^T\tilde{V}\|\leq \frac{4}{3}$ as shown in proof of theorem \ref{thm:final_err}.

\section{Complexity}\label{sec:complexity}

In this section, we will analyze the time complexity of each step in the main algorithm.\red{ We divide these steps into four parts and analyze each part in each subsection: Step 1-3 are considered in Subsection \ref{com1}. Step 4 is considered in Subsection \ref{com2}. Step 5-6 are considered in Subsection \ref{com3}. Step 7-8 are considered in Subsection \ref{sec:final_compute}.} Note that in the main algorithm the variables $R,\tilde{V}_l,\tilde{\alpha}$ are queried instead of calculated. We include the corresponding query complexity in analysis of the steps where we queried these variables.

\subsection{Sampling of columns and rows}\label{com1}
\red{In Step 1, the value of $r$ and $c$ are determined according to Inequalities (\ref{eq3},\ref{eq4},\ref{eq5},\ref{eq:r},\ref{eq:c}). The time of solving these inequalities is a constant. In Step 2 we sample $r$ indices, each sampling takes no more than $\log_2m$ time according to the arborescent vector data structure shown in \ref{subsection_sample}. In Step 3 we sample $c$ indices, each sampling takes no more than $r\log_2n$ time according to the arborescent matrix data structure shown in \ref{subsection_sample}. Thus the overall time complexity of Step 1-3 is} $O(r\log_2m+cr\log_2n)$.
\subsection{The spectral decomposition}\label{com2}
\red{Step 4 is the spectral decomposition.} For $r\times r$ symmetric matrix $A$, the fastest classical spectral decomposition is through classical spectral symmetric QR method, of which the complexity is $O(r^3)$.

\subsection{Calculation of $\tilde{\lambda_l}$}\label{com3}
\red{In Step 5-6 we calculate $\tilde{\lambda_l}$.} By Alg. \ref{estimation_alg}, we have
\begin{displaymath}
\lambda_l=\frac{1}{\sigma_l^2}V^{\prime\prime T}_lRy=\frac{1}{\sigma_l^2}\text{Tr}[V^{\prime\prime T}_lX^{\prime T}Xy]=\frac{1}{\sigma_l^2}\text{Tr}[XyV^{\prime\prime T}_lX^{\prime T}].
\end{displaymath}
Observe that
$\|yV^{\prime\prime T}_lX^{\prime T}\|_F=\|y\|\|V^{\prime\prime T}_lX^{\prime T}\|\leq \|y\|$, and we can query the $(i,j)$ matrix element of $yV^{\prime\prime T}_lX^{\prime T}$ in cost $O(r)$. According to Lemma \ref{lemma1}, the complexity in step \ref{step:elements} is
\begin{align*}
T_6=O(\frac{\|X\|_F^2k^2}{\epsilon^2}\log_2(\frac{8k}{\eta})(\log_2(mn)+k)).
\end{align*}

\subsection{Calculation of $x^TX\tilde{\alpha}$}\label{sec:final_compute}
\red{In Step 7-8 we calculate $x^TX\tilde{\alpha}$. }
Calculation of $x^TX\tilde{\alpha}$ is the last step of the algorithm, and also the most important step for saving time complexity. In Step \ref{step:find_sign} of Alg. \ref{main_alg}, we need to calculate $x^TX\tilde{\alpha}$, which is equal to $\text{Tr}[X\tilde{\alpha}x^T]$, with precision $\epsilon\|\alpha\|\|x\|$ and success probability $1-\frac{\eta}{4}$ using Alg. \ref{estimation_alg}. Let the $A$ and $B$ in Alg. \ref{estimation_alg} be $X$  and $\tilde{\alpha}x^T$, respectively. To calculate $\text{Tr}[X\tilde{\alpha}x^T]$, we first establish the query access for $\tilde{\alpha}x^T$ (we already have the sampling access of $X$), and then using the Alg. \ref{estimation_alg} as an oracle. \red{We first analyze the time complexity of querying $R$ and $\tilde{\alpha}$, and then provide the time complexity of calculating $x^TX\tilde{\alpha}$:}

\subsubsection{Query of $R$}
First we find query access of $R=X^{\prime T}X$. For any \black{$s=1,...,r,\quad j=1,...,m$}, $R_{sj}=e^T_sX^{\prime T}Xe_j=\text{Tr}[Xe_je^T_sX^{\prime T}]$, we calculate such trace by Alg. \ref{estimation_alg} to precision $\epsilon_1$ with success probability $1-\eta_1$. The time complexity for one query will be
\[Q(R)=O(\log_2(\frac{2}{\eta_1})\frac{\|X\|^4_F}{\epsilon_1^2r}\log_2(mn)).\]

\subsubsection{Query of $\tilde{\alpha}$}
For any $i=1,...,m$, we have $\tilde{\alpha}_j=\sum^r_{s=1}R_{sj}u_s$. One query of $\tilde{\alpha}$ will cost time $rkQ(R)$, with error $\epsilon_1\sum_{s=1}^r|u_s|$ and success probability more than $1-r\eta_1$.

\subsubsection{Calculation of $x^TX\tilde{\alpha}$}
We use Alg. \ref{estimation_alg} to calculate $x^TX\tilde{\alpha}=\text{Tr}[X\tilde{\alpha}x^T]$ to precision $\frac{\epsilon}{2}\|\alpha\|\|x\|$ with success probability $1-\frac{\eta}{8}$. Notice the query of $\tilde{\alpha}$ is with error and success probability. We only need
\[\epsilon_1\sum_{s=1}^r|u_s|\ceil{\frac{36}{\epsilon^2}}\ceil{6\log_2(\frac{16}{\eta})}\leq \frac{\epsilon}{2}\|\alpha\|\|x\|,\]
\[r\eta_1\ceil{\frac{36}{\epsilon^2}}\ceil{6\log_2(\frac{16}{\eta})}\leq\frac{\eta}{8}\]
to fulfill the overall computing task. Notice $\sum_{s=1}^r|u_s|\leq \sqrt{r}\|u\|$ and $\alpha=R^Tu$ We set
\begin{displaymath}
  \epsilon_1=\frac{\epsilon\|x\|}{2\sqrt{r}\ceil{\frac{36}{\epsilon^2}}\ceil{6\log_2(\frac{16}{\eta})}},
\end{displaymath}
  \begin{displaymath}
\eta_1=\frac{\eta}{8r\ceil{\frac{36}{\epsilon^2}}\ceil{6\log_2(\frac{16}{\eta})}}.
\end{displaymath}
And the overall time complexity for computing $x^TX\tilde{\alpha}$ is
\begin{align*}
  T_7&=O(\frac{1}{\epsilon^2}\log_2\frac{1}{\eta}(\log_2(mn)+rkQ(R)))\\
&=O(\frac{1}{\epsilon^2}\log_2\frac{1}{\eta}(\log_2(mn)+rk\log_2(\frac{2}{\eta_1})\frac{\|X\|^4_F}{\epsilon_1^2r}\log_2(mn))).
\end{align*}

\section{Experiments}
In this section, we demonstrate the proposed quantum-inspired SVM algorithm in practice by testing the algorithm on artificial datasets. The feasibility and efficiency of some other quantum-inspired algorithms (quantum-inspired algorithms for recommendation systems and linear systems of equations) on large datasets has been benchmarked, and the results indicate that quantum-inspired algorithms can perform well in practice under its specific condition: low rank, low condition number,
and very large dimension of the input matrix \cite{qip}.
Here we show the feasibility of the quantum-inspired SVM. Firstly, we test the quantum-inspired SVM algorithm on low-rank and low-rank approximated datasets and compare it to an existing classical SVM implementation. Secondly, we discuss the characteristics of the algorithm by analyzing its dependence on the parameters and datasets. In our experiment, we use the arborescent data structure instead of arrays for storage and sampling \cite{qip}, making the experiment conducted in a more real scenario compared to the previous work \cite{qip}. All algorithms are implemented in Julia \cite{bezanson2017julia}. The source code and data are available at \url{https://github.com/helloinrm/qisvm}.

\subsection{Experiment I: Comparison with LIBSVM}

In this experiment, we test quantum-inspired SVM algorithm on large datasets and compare its performance to the well-known classical SVM implementation LIBSVM \cite{CC01a}.

We generate datasets of size $10000\times 11000$, which represent 11000 vectors (6000 vectors for training and 5000 vectors for testing) with length 10000. All the data vectors \black{in training and testing sets are chosen uniformly at random} from the generated data matrix, so that they are statistically independent and \black{identically distributed}. We test quantum-inspired SVM and LIBSVM on two kinds of datasets: low-rank datasets (rank$=1$) and high-rank but low-rank approximated datasets (rank$=10000$). Each scenario is repeated for 5 times. The construction method for data matrices is described in Appendix~\ref{app:generate}. And the parameters for quantum-inspired SVM are choosen as $\epsilon=5,\eta=0.1$ and $b=1$ (We explain the parameters and their setting in Experiment II.).

The average classification rates are shown in Table~\ref{table}, from which we observe the advantage of quantum-inspired SVM on such low-rank approximated datasets (on average about $5\%$ higher). We also find that both quantum-inspired SVM and LIBSVM performs better on low-rank datasets than low-rank approximated datasets.

\begin{table*}[!t]
  \caption{The average values and standard deviations of classification rates (\%) of qiSVM and LIBSVM in five experiments.}\label{table}
  \renewcommand{\arraystretch}{1.3}
  \centering
  \begin{tabular}{c||c|c|c|c}
  \hline
  & \multicolumn{2}{c|}{\bfseries Testing Set} & \multicolumn{2}{c}{\bfseries Training Set}\\
  \hline
  &qiSVM&LIBSVM&qiSVM&LIBSVM\\
  \hline
  Low-rank &91.45$\pm$3.17 & 86.46$\pm$2.00&91.35$\pm$3.64&86.45$\pm$2.15\\
  \hline
  Low-rank approximated &89.82$\pm$4.38&84.90$\pm$3.20&89.92$\pm$4.23&84.69$\pm$2.87\\
  \hline
  \end{tabular}
  \end{table*}

  \begin{figure*}[!t]
    \centering
    \subfloat[]{\includegraphics[width=0.3\textwidth]{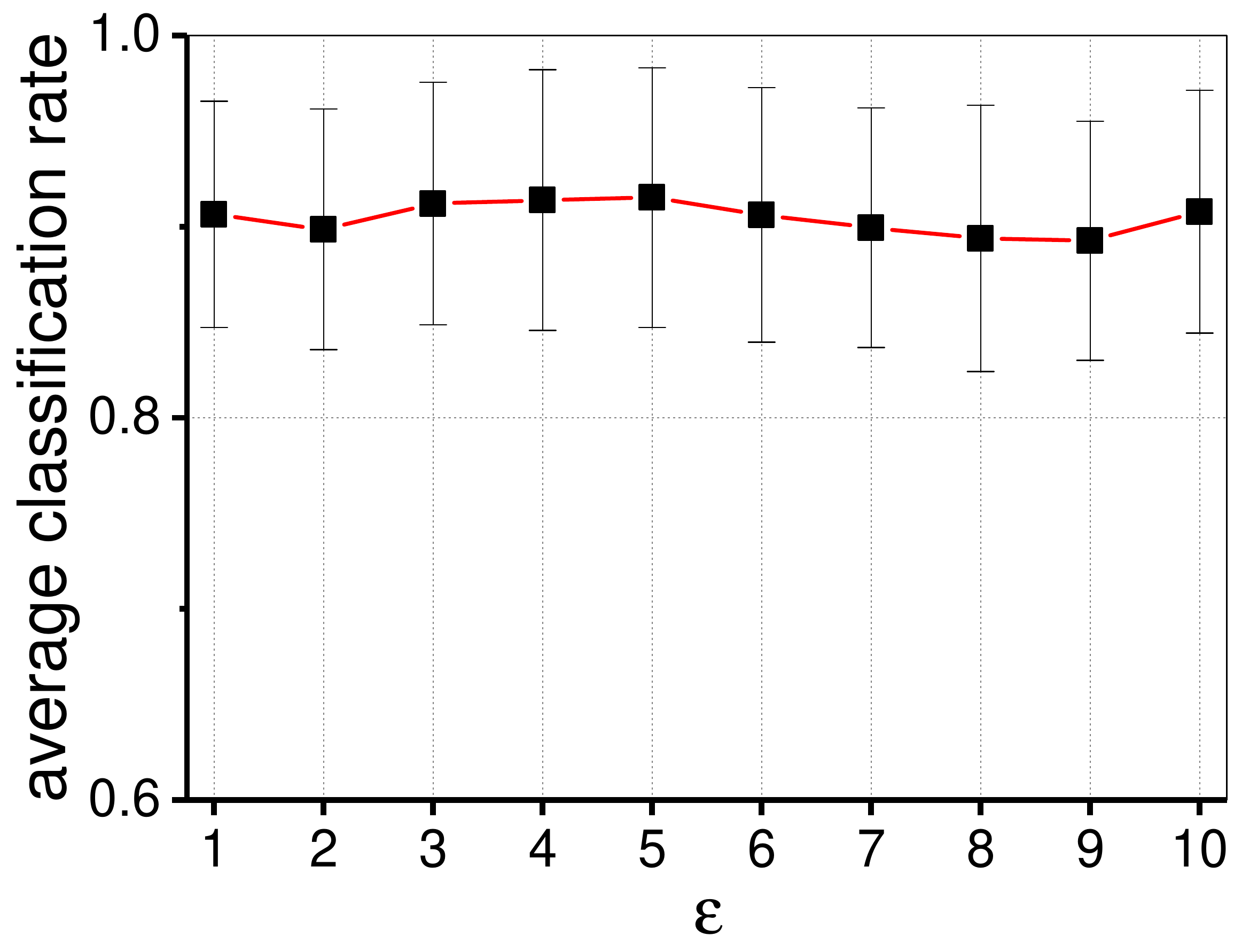}
    \label{epsilon}}
    \subfloat[]{\includegraphics[width=0.3\textwidth]{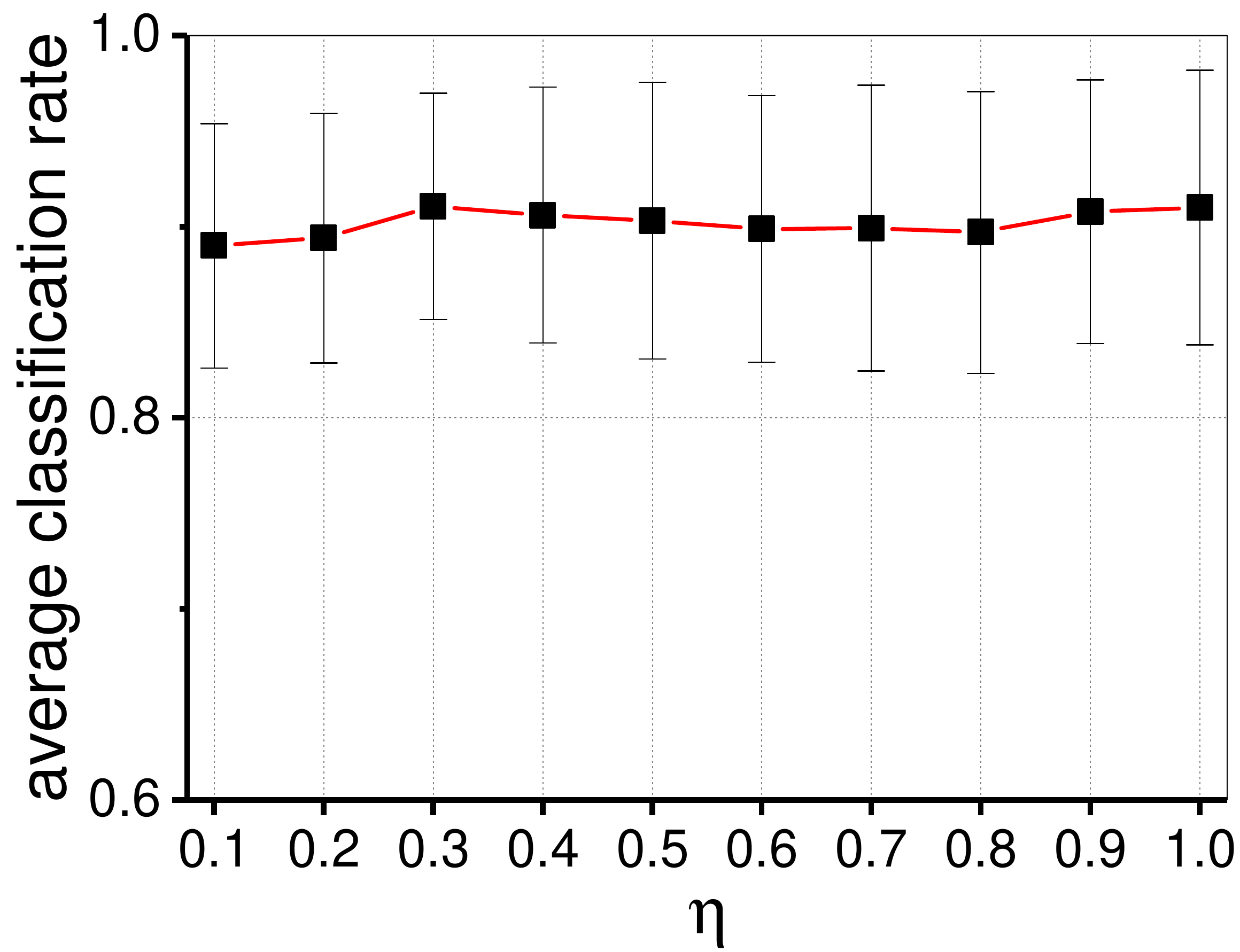}
    \label{eta}}
    \subfloat[]{\includegraphics[width=0.3\textwidth]{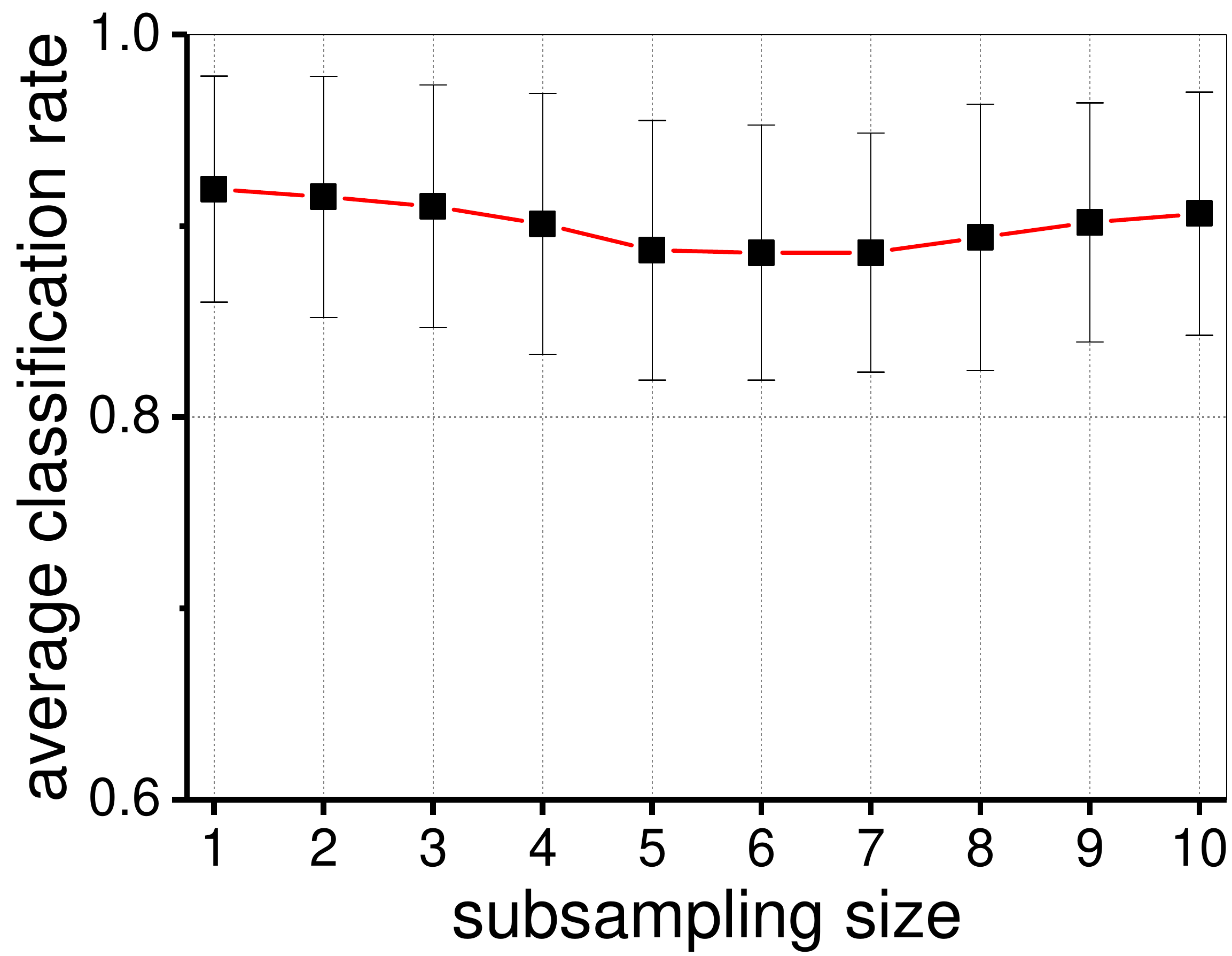}
    \label{ss}}
    \caption{The average classification rate of quantum-inspired SVM algorithm with different parameters on the dataset with rank 1. Each point represents an average classification rate for 50 trials, and the error bar shows the standard deviation of the 50 trials. (a) Algorithm performance when the parameter $\epsilon$ is taken from 1 to 10. (b) Algorithm performance when the parameter $\eta$ is taken from 0.1 to 1. (c) Algorithm performance when the parameter $b$ is taken from 1 to 10.}
    \label{fig:parameters}
  \end{figure*}

\subsection{Experiment II: Discussion on algorithm parameters}

As analyzed in Section~\ref{sec:accuracy} and Section~\ref{sec:complexity}, there are two main parameters for the quantum-inspired algorithm: relative error $\epsilon$ and success probability $1-\eta$. Based on them we set subsampling size $r,c$ and run the algorithm. However, for datasets that are not large enough, setting $r,c$ by Equation~(\ref{eq:r}) and Equation~(\ref{eq:c}) is rather time costly. For instance, when the condition number of data matrix is 1.0, taking $\eta=0.1$ and $\epsilon=5.0$, theoretically, the $r,c$ for $10000\times 10000$ dataset should be set as 1656 and 259973 to assure that the algorithm calculates the classification expression with relative error less than $\epsilon$ and success probability higher than $1-\eta$. For practical applications of not too large datasets, we set $r,c$ as $r=b\ceil{4\log_2(2n/\eta)/\epsilon^2}$ and $c=b\ceil{4\log_2(2r/\eta)/\epsilon^2}$, in which $b$ is the subsampling size control parameter. When $b=1$, our practical choice of $r,c$ assures the relative error of subsampling (Step~\ref{step:sample_cols} and Step~\ref{step:sample_rows} in Alg.~\ref{main_alg}) won't exceed $\epsilon$ (guaranteed by Theorem~\ref{theorem:unbiased_matrix}).

In Experiment I, we took the practical setting of $r,c$, where we already found advantage compared to LIBSVM. Our choice of $\epsilon, \eta$ and $b$ is $\epsilon=5$, $\eta=0.1$ and $b=1$. Here, we test the algorithm on other choices of $\epsilon, \eta$ and $b$ and check the classification rate of the algorithm. We test each parameter choice for 50 times. The variation intervals of each parameter are $\epsilon$ from 1 to 10, $\eta$ from 0.1 to 1, and $b$ from 1 to 10. The results are shown in Fig.~\ref{fig:parameters}. We find the average classification rates of the algorithm in each experiment are close. We notice when using the practical $r,c$, which are much smaller than the theoretical ones, the algorithm maintains its performance (classification rate around 0.90). This phenomenon indicates a gap between our theoretical analysis and the actual performance, as \cite{qip} reports ``the performance of these algorithms is better than the theoretical
complexity bounds would suggest''.

\section{Discussion}\label{sec:discuss}

In this section, we will present some discussions on the proposed algorithm. And we will also discuss the potential applications of our techniques to other types of SVMs, such as non-linear SVM and least square SVM, but more works on the complexity and errors are required in future work if we want to realize these extensions.

\subsection{The cause of exponential speedup}
An interesting fact is that we can achieve exponential speedup without using any quantum resources, such as superposition or entanglement. This is a somewhat confusing but reasonable result that can be understood as follows: Firstly, the advantage of quantum algorithms, such as HHL algorithm, is that high-dimensional vectors can be represented using only a few qubits. By replacing qRAM to the arborescent data structure for sampling, we can also represent the low-rank matrices by its normalized submatrix in a short time. By using the technique of sampling, large-size calculations are avoided, and we only need to deal with the problem that has the logarithmic size of the original data. Secondly, the relative error of matrix subsampling algorithm is minus-exponential on the matrix size, which ensures the effectiveness of such logarithmic-complexity algorithm (e.g. Theorem 2 shows the error of matrix row subsampling).

\subsection{Improving sampling for dot product}
Remember in Alg. \ref{estimation_alg} we can estimate dot products for two vectors. However, it does not work well for all the conditions, like when $\|x\|$ and $\|y\|$ are donminated by one element. For randomness, \cite{sampling_tech} implies that we can apply a spherically random rotation $R$ to all $x$, which does not change the kernel matrix $K$, but will make all the elements in the dataset matrix be in a same distribution.

\subsection{LS-SVM with non-linear kernels}\label{subsec:non-linear-svm}
In Section \ref{sec:pre}, we have considered the LS-SVM with the linear kernel $K=X^TX$. When data sets are not linear separable, non-linear kernels are usually needed. To deal with non-linear kernels with Alg.~\ref{main_alg}, we only have to show how to establish sampling access for the non-linear kernel matrix $K$ from the sampling access of $X$.

We first show how the sampling access of polynomial kernel
$K_p(x_i,x_j)=(x_j^Tx_i)^p$ can be established. The corresponding kernel matrix is $K_p=((x_j^Tx_i)^p)_{i=1,\dots,m,j=1,\dots,m}$.

We take
\begin{displaymath}
  Z=(x_1^{\otimes p},x_2^{\otimes p},...,x_m^{\otimes p}),
\end{displaymath}
in which the $j$-column $Z_j$ is the $p$-th tensor power of $x_j$.

Notice that $Z^TZ=K_p$. Once we have sampling access of $Z$, we can sample $K_p$ as Step~\ref{step:sample_cols} and Step~\ref{step:sample_rows} in Alg.~\ref{main_alg} do.
The sampling access of $Z$ can be established by (The effectiveness of Alg.~\ref{alg-K-sampling} is shown in  Appendix~\ref{app:alg4-proof}.):

\begin{algorithm}[H]
  \centering
  \begin{algorithmic}[1]
  \item[\textbf{Input:}] The sampling access of $X$ in logarithmic time of $m$ and $n$.
  \item[\textbf{Goal:}] Sample a column index $j$ from the column norm vector $(\|x_1\|^{p},\|x_2\|^{p},\dots,\|x_m\|^{p})$ of $Z$, and them sample a row index $i$ from column $x_j^{\otimes p}$ of $Z$.
  \State Sample on column norm vector $(\|x_1\|,\|x_2\|,\dots,\|x_m\|)$ of $X$ to get index $j$.\label{step:K-sam-1}
  \State Query $\|x_j\|$ from $(\|x_1\|,\|x_2\|,\dots,\|x_m\|)$. Calculate $\|x_j\|^p$.
  \State Sample a real number $a$ uniformly distributed in $[0,1]$. If $a\geq\|x_j\|^p$, go to Step \ref{step:K-sam-1}. If not, output index $j$ as the column index and continue.
  \algstore{bkbreak}
\end{algorithmic}
\caption{Polynomial kernel matrices sampling.}\label{alg-K-sampling}
\end{algorithm}
\begin{algorithm}[H]
  \addtocounter{algorithm}{-1}
\centering
\begin{algorithmic}[1]
\algrestore{bkbreak}
  \State Repeat sampling on $x_j$ for $p$ times. Denote the outcome indices as $i_1,i_2,\dots,i_p$.
  \item[\textbf{Output:}] Column index $j$ and row index $\sum_{\tau=1}^p(i_\tau-1)n^{p-\tau}+1$.
  \end{algorithmic}
  \caption{Polynomial kernel matrices sampling.}\label{alg-K-sampling}
\end{algorithm}

\black{
For general non-linear kernels, we note that they can always be approximated by linear combination of polynomial kernels (and thus can be sampled based on sampling access of polynomial kernels) the corresponding non-linear feature function is continous. For instance, the popular radial basis function (RBF) kernel
\[K_{\text{RBF}}(x_i,x_j)=\text{exp}(-\frac{\|x_i-x_j\|^2}{2\sigma^2})\]
can be approximated by
\begin{align*}
\tilde{K}_{\text{RBF}}(x_i,x_j)&=\sum^N_{p=0}\frac{1}{p!}\left(-\frac{x_i^Tx_i-2x_j^Tx_i+x^T_jx_j}{2\sigma^2}\right)^p\\
&=\sum^N_{p=0}\left(-\frac{1}{2\sigma^2}\right)^p\sum^p_{q,l=0}\binom{p}{q}K_q(x_i,x_i)\\
&+(-2)^l\binom{p}{l}K_l(x_i,x_j)+\binom{p}{p-q-l}K_{p-q-l}(x_j,x_j).
\end{align*}
}

\subsection{General LS-SVM}\label{subsec:lssvm}
\black{In the former sections, we began with a LS-SVM with $b=0$ and linear kernels in Section \ref{sec:pre}. And we showed how the method can be extended to nonlinear kernels in Section \ref{subsec:non-linear-svm}. Finally, we deal with the last assumption $b=0$. We show how a general LS-SVM can be tackled using techniques alike in Alg.~3:}

A general LS-SVM equation\cite{lssvm} is
\begin{gather}\label{eqn:lssvm}
  \begin{pmatrix}
    0&1^T\\1&K+\gamma^{-1}I
    \end{pmatrix}\begin{pmatrix}
      b\\\alpha
    \end{pmatrix}=\begin{pmatrix}
      0\\y
    \end{pmatrix},
\end{gather}
in which $K$ is the kernel matrix.

Equation~(\ref{eqn:lssvm}) can be solved as follows:

(i) Firstly, by methods in Section \ref{subsec:non-linear-svm}, we establish the sampling access for kernel matrix $K$. Suppose a sampling outcome of $K$ is $K''$.

(ii) Secondly, take \[A=\begin{pmatrix}
  0&1^T\\1&K+\gamma^{-1}I
  \end{pmatrix}.\]
  and
  \[A''=\begin{pmatrix}
    0&1^T\\1&K''+\gamma^{-1}I
    \end{pmatrix}.\]
We establish the eigen relations between $A$ and $A''$ by theorems which are similar to Theorem~\ref{theorem:unbiased_matrix} and Theorem~\ref{thm:to_prove}.

(iii) Once $A\in \mathbb{R}^{m\times m}$ is subsampled to $A''\in \mathbb{R}^{r\times r}$, we can continue Step \ref{step:sample_rows}-\ref{step:find_query} of Alg.~\ref{main_alg}.

(iv) \black{Once Equation~(\ref{eqn:lssvm}) is solved in Step \ref{step:find_query} of Alg.~\ref{main_alg}, which means we can establish the query access for $\alpha$. According to Equation \ref{eqn:lssvm}, $b=y_j-x_j^TX\alpha-\gamma^{-1}\alpha_j$ for any $j$ such that $\alpha_j\neq 0$. We can then evaluate the classification expression $y_j+(x-x_j)^TX\alpha-\gamma^{-1}\alpha_j$ and make classification using Alg.~\ref{estimation_alg}. There are two ways to find $j$: One is executing the rejection sampling on $\alpha$ using Alg. \ref{rejection_alg}. The other is checking if $\alpha_j=0$ after each sampling of $X$ in Step 3 of Alg. \ref{estimation_alg}.
}

\section{Conclusion}
We have proposed a quantum-inspired SVM algorithm that achieves exponential speedup over the previous classical algorithms. The feasibility of the proposed algorithm is demonstrated by experiments. Our algorithm works well on low-rank datasets or datasets that can be well approximated by low-rank matrices, which is similar with quantum SVM algorithm\cite{qsvm} as "when a low-rank approximation is appropriate". Certain investigations on the application of such an algorithm are required to make quantum-inspired SVM operable in solving questions like face recognition\cite{face} and signal processing\cite{book2}.

We hope that the techniques developed in our work can lead to the emergence of more efficient classical algorithms, such as applying our method to support vector machines with more complex kernels \cite{lssvm,book} or other machine learning algorithms.
The technique of indirect sampling can expand the application area of fast sampling techniques. And it will make contribution to the further competition between classical algorithms and quantum ones.


Some improvements on our work would be made in the future, such as reducing the conditions on the data matrix, further reducing the complexity, and tighten the error bounds in the theoretical analysis, which can be achieved through a deeper investigation on the algorithm and the error propagation process. The investigation of quantum-inspired non-linear SVMs and least squares SVM as discussed in Section \ref{sec:discuss} also requires theoretical analysis and empirical evaluations.

We note that our work, as well as the previous quantum-inspired algorithms, are not intended to
demonstrate that quantum computing is uncompetitive.
We want to find out where the boundaries of
classical and quantum computing are, and we expect new quantum algorithms to be developed to beat
our algorithm.

\appendices

\section{Proof of Theorems in \ref{sec:accuracy}}\label{app:proof}
\subsection{Proof of Theorem \ref{thm1}}
\begin{IEEEproof}
\red{We break the expression $|V_i^{\prime \prime T}A'V''_j-\delta_{ij}\sigma_i^2|$ into two parts,}
\[|V_i^{\prime \prime T}A'V''_j-\delta_{ij}\sigma_i^2|\leq |V_i^{\prime \prime T}(A'-A'')V''_j|+|V_i^{\prime \prime T}A''V''_j-\delta_{ij}\sigma_i^2|.\]
\red{For the first item, because of the condition $\|A'-A''\|\leq\beta$ and $V''_j$ are normalized,}
\begin{align*}
  |V_i^{\prime \prime T}(A'-A'')V''_j| & \leq \|V_i^{\prime \prime T}\|\cdot\|(A'-A'')V''_j\|\\
  & \leq \beta.
  \end{align*}
  \red{For the second item, because of the condition $A''=\sum^k_{l=1}\sigma^2_lV''_lV^{\prime\prime T}_l$,}
\[|V_i^{\prime \prime T}A''V''_j-\delta_{ij}\sigma_i^2|=0.\]
\red{In all,}
\[|V_i^{\prime \prime T}A'V''_j-\delta_{ij}\sigma_i^2|\leq\beta.\]
\red{The description above can be written in short as follows:}
\begin{align*}
  |V_i^{\prime \prime T}A'V''_j-\delta_{ij}\sigma_i^2|&\leq |V_i^{\prime \prime T}(A'-A'')V''_j|+|V_i^{\prime \prime T}A''V''_j-\delta_{ij}\sigma_i^2| \\
  & \leq \|V_i^{\prime \prime T}\|\cdot\|(A'-A'')V''_j\|\\
  & \leq \beta.
  \end{align*}
\end{IEEEproof}

\subsection{Proof of Theorem \ref{thm:to_prove}}
\begin{IEEEproof}
\red{Denote $|\tilde{V}_i^T\tilde{V}_j-\delta_{ij}|$ as $\Delta_1$, $|\tilde{V}_i^TA\tilde{V}_j-\delta_{ij}\sigma^2_i|$ as $\Delta_2$. By definition, $\tilde{V}_l=\frac{1}{\sigma^2_l}R^TV''_l$. Thus}
\[\Delta_1=|\frac{V^{\prime\prime T}_iRR^TV''_j-\delta_{ij}\sigma^4_i}{\sigma_i^2\sigma_j^2}|.\]
\red{We break it into two parts:}
\[\Delta_1\leq\frac{1}{\sigma_i^2\sigma_j^2}\left(|V^{\prime\prime T}_iA'A'V''_j-\delta_{ij}\sigma^4_i|+|V^{\prime\prime T}_i(RR^T-A'A')V''_j|\right).\]
\red{For the first item, we have}
\begin{align*}
  &|V^{\prime\prime T}_iA'A'V''_j-\delta_{ij}\sigma^4_i|\\
  =&|V^{\prime\prime T}_i(A'-A'')^2V''_j+V^{\prime\prime T}_i(A'-A'')A'V''_j\\
  +&V^{\prime\prime T}_iA'(A'-A'')V''_j+V^{\prime\prime T}_iA''A''V''_j-\delta_{ij}\sigma^4_i|\\
  \leq&|V^{\prime\prime T}_i(A'-A'')^2V''_j|+|V^{\prime\prime T}_i(A'-A'')A'V''_j|\\
  +&|V^{\prime\prime T}_iA'(A'-A'')V''_j|+|V^{\prime\prime T}_iA''A''V''_j-\delta_{ij}\sigma^4_i|\\
  \leq&\beta^2+\sigma^2_j\beta+\sigma^2_i\beta.
\end{align*}
\red{The last step above used the same technique as the proof of Thm 3.}

\red{For the second item, we have}
\begin{align*}
  |V^{\prime\prime T}_i(RR^T-A'A')V''_j|&\leq\|RR^T-A'A'\|\\
&=\|X^{\prime T}XX^TX'-X^{\prime T}X'X^{\prime T}X'\|\\
&\leq\|X'\|^2\|XX^T-X'X^{\prime T}\|.
\end{align*}
\red{Because}
\[\|X'\|\leq\|X'\|_F=\|X\|_F,\]
\red{we have}
\[|V^{\prime\prime T}_i(RR^T-A'A')V''_j|\leq\epsilon'\|X\|_F^2.\]
\red{In all, due to $\sigma_i\geq\kappa\quad\forall i\in\{1,\dots,k\}$,}
\begin{align*}
  \Delta_1&\leq\frac{1}{\sigma_i^2\sigma_j^2}(\beta^2+\sigma^2_j\beta+\sigma^2_i\beta+\epsilon'\|X\|_F^2)\\
  &\leq\kappa^2\beta^2+2\kappa\beta+\kappa^2\epsilon'\|X\|^2_F.
\end{align*}
\red{By definition, $\tilde{V}_l=\frac{1}{\sigma^2_l}R^TV''_l$. Thus}
\[\Delta_2=|\frac{V^{\prime\prime T}_iRAR^TV''_j-\delta_{ij}\sigma^6_i}{\sigma_i^2\sigma_j^2}|.\]
\red{We break it into two parts:}
\begin{align*}
  \Delta_2&\leq\frac{1}{\sigma_i^2\sigma_j^2}(|V_i^{\prime \prime T}(RAR^T-A'A'A')V''_j|\\
  &+|V_i^{\prime \prime T}A'A'A'V''_j-\delta_{ij}\sigma_i^6|)
\end{align*}
\red{For the first item, we have}
\begin{align*}
 &\quad |V_i^{\prime \prime T}(RAR^T-A'A'A')V''_j|\\
  &\leq\|RAR^T-A'A'A'\|\\
  &\leq\|X'\|^2\|XX^TXX^T-X'X^{\prime T}X'X^{\prime T}\|\\
  &\leq\|X\|^2_F(\|XX^T(XX^T-X'X^{\prime T})\|+\|(XX^T-X'X^{\prime T})X'X^{\prime T}\|)\\
  &\leq 2\|X\|^2_F\|X\|^2\|XX^T-X'X^{\prime T}\|\\
  &\leq 2\|X\|^2_F\epsilon'.
\end{align*}
\red{For the second item, we have}
\begin{align*}
  &|V_i^{\prime \prime T}A'A'A'V''_j-\delta_{ij}\sigma_i^6|\\
  =&|V_i^{\prime \prime T}(A'-A'')A'A'V''_j+V_i^{\prime \prime T}A''(A'-A'')A'V''_j\\
  +&V_i^{\prime \prime T}A''A''(A'-A'')V''_j+V_i^{\prime \prime T}A''A''A''V''_j-\delta_{ij}\sigma_i^6|\\
  \leq&|V_i^{\prime \prime T}(A'-A'')A'A'V''_j|+|V_i^{\prime \prime T}A''(A'-A'')A'V''_j|\\
  +&|V_i^{\prime \prime T}A''A''(A'-A'')V''_j|+|V_i^{\prime \prime T}A''A''A''V''_j-\delta_{ij}\sigma_i^6|\\
  \leq&\|(A'-A'')A'A'\|+\|A''(A'-A'')A'\|+\|A''A''(A'-A'')\|\\
  \leq&\|X'\|^4\|A'-A''\|+\|X''\|^2\|X'\|^2\|A'-A''\|\\
  +&\|X''\|^4\|A'-A''\|\\
  \leq&\beta\|X\|^4_F.
\end{align*}
\red{In all,}
\begin{align*}
  \Delta_2&\leq\frac{1}{\sigma_i^2\sigma_j^2}(2\|X\|^2_F\epsilon'+\beta\|X\|^4_F)\\
  &\leq (2\epsilon'+\beta\|X\|^2_F)\|X\|^2_F\kappa^2.
\end{align*}
\end{IEEEproof}
\subsection{Proof of Theorem \ref{thm:final_err}}
\begin{IEEEproof}
\red{For $\tilde{V}_i^T\tilde{V}_j-\delta_{ij}$ are elements of $\tilde{V}^T\tilde{V}-I$ and $|\tilde{V}_i^T\tilde{V}_j-\delta_{ij}|\leq \frac{1}{4k}$,}
\[\|\tilde{V}^T\tilde{V}-I\|\leq k \text{max}\{|\tilde{V}_i^T\tilde{V}_j-\delta_{ij}|\}\leq \frac{1}{4}.\]
\red{Thus $\|\tilde{V}^T\tilde{V}\|$ is invertible and}
\[\|(\tilde{V}^T\tilde{V})^{-1}\|=1/\|\tilde{V}^T\tilde{V}\|\leq 1/(1-\|\tilde{V}^T\tilde{V}-I\|)=\frac{4}{3}.\]
\red{Take $B=\tilde{V}\Sigma^{-2}\tilde{V}^TA-I_m$, we have}
\[|\tilde{V}_i^TB\tilde{V}_j|=|\sum^k_{l=1}\frac{\tilde{V}_i^T\tilde{V}_l\cdot \tilde{V}_l^TA\tilde{V}_j}{\sigma_l^2}-\tilde{V}^T_i\tilde{V}_j|.\]
\red{We break it into two parts:}
\[|\tilde{V}_i^TB\tilde{V}_j|\leq|\sum^k_{l=1}\frac{\tilde{V}^T_i\tilde{V}_l}{\sigma_l^2}(\tilde{V}_l^TA\tilde{V}_j-\delta_{lj}\sigma^2_l)|+|\sum^k_{l=1}\tilde{V}^T_i\tilde{V}_l\delta_{lj}-\tilde{V}^T_i\tilde{V}_j|.\]
\red{The second item is zero because}
\[|\sum^k_{l=1}\tilde{V}^T_i\tilde{V}_l\delta_{lj}-\tilde{V}^T_i\tilde{V}_j|=|\tilde{V}^T_i\tilde{V}_j-\tilde{V}^T_i\tilde{V}_j|.\]
\red{The first item}
\begin{align*}
  |\sum^k_{l=1}\frac{\tilde{V}^T_i\tilde{V}_l}{\sigma_l^2}(\tilde{V}_l^TA\tilde{V}_j-\delta_{lj}\sigma^2_l)|&\leq\zeta\kappa|\sum^k_{l=1}\tilde{V}^T_i\tilde{V}_l|\\
  &\leq \zeta\kappa(\sum_{l\neq i}|\tilde{V}^T_i\tilde{V}_l|+|\tilde{V}^T_i\tilde{V}_i|)\\
  &\leq \zeta\kappa((k-1)\frac{1}{4k}+(\frac{1}{4k}+1))\\
  &\leq \frac{5}{4}\zeta\kappa.
  \end{align*}
  \red{Thus $|\tilde{V}_i^TB\tilde{V}_j|\leq \frac{5}{4}\zeta\kappa$ and $\|\tilde{V}^TB\tilde{V}\|\leq \frac{5}{4}\zeta\kappa k$.  By Theorem \ref{thm:err_of_B},}
\[\|\tilde{V}\Sigma^{-2}\tilde{V}^TA-I_m\|=\|B\|\leq\|(\tilde{V}^T\tilde{V})^{-1}\|\|\tilde{V}^TB\tilde{V}\|\leq \frac{5}{3}\kappa k\zeta.\]


\end{IEEEproof}

\section{The construction method of datasets}\label{app:generate}
In our experiment, we constructed artificial datasets which are low-rank or can be low-rank approximated. Here we put up our construction mehtod:

1. Firstly, we \black{multiply} a random matrix $A$ of size $n\times k$ with another random matrix $B$ of size $k\times m$. The elements in both of them are evenly distributed in $[-0.5,0.5]$. Denote the multiplication outcome as $X$. Then the rank of $X$ is at most $k$.

2. We add turbulence to the matrix $X$ by adding a random number evenly distributed in $[-0.1x,0.1x]$ to all the elements in $X$, in which $x$ is the average of all the absolute values of $X$. After adding turbulence, $X$ is no more low-rank but still low-rank approximated.

3. We normalize $X$ such that $X$ has operator norm 1.

4. We divide the column vectors of $X$ into two classes by a random hyperplane \black{$w^Tx=0$ that passes the origin (By random hyperplane we mean the elements in $w$ are uniformly sampled from $[0,1]$ at random.)}, while making sure that both classes are not empty.

5. Since now we have $m$ linear-separable labeled vectors, each with length $n$. We \black{choose uniformly at random} $m_1$ of them for training, and let the other $m_2=m-m_1$ for testing, while making sure that the training set includes vectors of both classes.

\section{The effectiveness of Alg.~\ref{alg-K-sampling}}\label{app:alg4-proof}

The goal of Alg.~\ref{alg-K-sampling} is to sample a column index and a row index from $Z$. We show it achieves this goal.

Step 1-3 are for sampling out the column index. They are essentially Alg.~\ref{rejection_alg} with $A=\text{Diag}(\|x_1\|^{p-1},\dots,\|x_m\|^{p-1})$ and $b=(\|x_1\|,\dots,\|x_m\|)$, which sample from the column norm vector $b=(\|x_1\|^p,\dots,\|x_m\|^p)$  of $Z$ to get the column index $j$. We note that in practical applications, Step 1-3 can be adjusted for speedup, such as frugal rejection sampling suggested in \cite{frugal_2018}.

Step 4 is for sampling out the row index. Suppose $l=\sum^p_{\tau=1}(i_\tau-1)n^{p-\tau}+1$. According the definition of tensor power, the $l$-th element of $x_j^{\otimes p}$ is
\[(x_j^{\otimes p})_l=\Pi^p_{\tau=1}x_{i_\tau j}.\]
When Step 4 executes $p$ times of sampling on $x_j$, the probability of getting the outcome $i_1,i_2,\dots,i_p$ is $|\Pi^p_{\tau=1}x_{i_\tau j}|^2$, which is exactly the probability of sampling out $(x_j^{\otimes p})_l$ in $x_j^{\otimes p}$. Thus we output index $l=\sum^p_{\tau=1}(i_\tau-1)n^{p-\tau}+1$.


\section*{Acknowledgment}
The authors would like to thank Yi-Fei Lu for helpful discussions.

\ifCLASSOPTIONcaptionsoff
\newpage
\fi

\bibliographystyle{IEEEtran}
\bibliography{b}

\begin{thebibliography}{10}
\providecommand{\url}[1]{#1}
\csname url@samestyle\endcsname
\providecommand{\newblock}{\relax}
\providecommand{\bibinfo}[2]{#2}
\providecommand{\BIBentrySTDinterwordspacing}{\spaceskip=0pt\relax}
\providecommand{\BIBentryALTinterwordstretchfactor}{4}
\providecommand{\BIBentryALTinterwordspacing}{\spaceskip=\fontdimen2\font plus
\BIBentryALTinterwordstretchfactor\fontdimen3\font minus
  \fontdimen4\font\relax}
\providecommand{\BIBforeignlanguage}[2]{{%
\expandafter\ifx\csname l@#1\endcsname\relax
\typeout{** WARNING: IEEEtran.bst: No hyphenation pattern has been}%
\typeout{** loaded for the language `#1'. Using the pattern for}%
\typeout{** the default language instead.}%
\else
\language=\csname l@#1\endcsname
\fi
#2}}
\providecommand{\BIBdecl}{\relax}
\BIBdecl

\bibitem{huang2020superconducting}
H.-L. Huang, D.~Wu, D.~Fan, and X.~Zhu, ``Superconducting quantum computing: a
  review,'' \emph{Science China Information Sciences}, vol.~63, no. 180501,
  2020.

\bibitem{shor1994algorithms}
\BIBentryALTinterwordspacing
P.~W. Shor, ``Algorithms for quantum computation: Discrete logarithms and
  factoring,'' in \emph{Proc. 35th Annual Symposium Foundations Computer
  Sci.}\hskip 1em plus 0.5em minus 0.4em\relax Santa Fe, NM, USA: IEEE, Nov.
  1994, pp. 124--134. [Online]. Available:
  \url{https://ieeexplore.ieee.org/document/365700}
\BIBentrySTDinterwordspacing

\bibitem{lu2007demonstration}
\BIBentryALTinterwordspacing
C.-Y. Lu, D.~E. Browne, T.~Yang, and J.-W. Pan, ``Demonstration of a compiled
  version of shor's quantum factoring algorithm using photonic qubits,''
  \emph{Physical Review Letters}, vol.~99, no.~25, p. 250504, 2007. [Online].
  Available:
  \url{https://journals.aps.org/prl/abstract/10.1103/PhysRevLett.99.250504}
\BIBentrySTDinterwordspacing

\bibitem{huang2017experimental}
\BIBentryALTinterwordspacing
H.-L. Huang, Q.~Zhao, X.~Ma, C.~Liu, Z.-E. Su, X.-L. Wang, L.~Li, N.-L. Liu,
  B.~C. Sanders, C.-Y. Lu \emph{et~al.}, ``Experimental blind quantum computing
  for a classical client,'' \emph{Physical review letters}, vol. 119, no.~5, p.
  050503, 2017. [Online]. Available:
  \url{https://journals.aps.org/prl/abstract/10.1103/PhysRevLett.119.050503}
\BIBentrySTDinterwordspacing

\bibitem{grover1996fast}
\BIBentryALTinterwordspacing
L.~K. Grover, ``A fast quantum mechanical algorithm for database search,'' in
  \emph{Proc. 21th Annual ACM Symposium Theory Computing}.\hskip 1em plus 0.5em
  minus 0.4em\relax Philadelphia, Pennsylvania, USA: ACM, May 1996, pp.
  212--219. [Online]. Available: \url{http://doi.acm.org/10.1145/237814.237866}
\BIBentrySTDinterwordspacing

\bibitem{li2018complementary}
\BIBentryALTinterwordspacing
T.~Li, W.-S. Bao, H.-L. Huang, F.-G. Li, X.-Q. Fu, S.~Zhang, C.~Guo, Y.-T. Du,
  X.~Wang, and J.~Lin, ``Complementary-multiphase quantum search for all
  numbers of target items,'' \emph{Physical Review A}, vol.~98, no.~6, p.
  062308, 2018. [Online]. Available:
  \url{https://journals.aps.org/pra/abstract/10.1103/PhysRevA.98.062308}
\BIBentrySTDinterwordspacing

\bibitem{qml}
\BIBentryALTinterwordspacing
J.~Biamonte, P.~Wittek, N.~Pancotti, P.~Rebentrost, N.~Wiebe, and S.~Lloyd,
  ``Quantum machine learning,'' \emph{Nature}, vol. 549, no. 7671, p.
  195–202, Sept. 2017. [Online]. Available:
  \url{https://doi.org/10.1038/nature23474}
\BIBentrySTDinterwordspacing

\bibitem{huang2018demonstration}
\BIBentryALTinterwordspacing
H.-L. Huang, X.-L. Wang, P.~P. Rohde, Y.-H. Luo, Y.-W. Zhao, C.~Liu, L.~Li,
  N.-L. Liu, C.-Y. Lu, and J.-W. Pan, ``Demonstration of topological data
  analysis on a quantum processor,'' \emph{Optica}, vol.~5, no.~2, pp.
  193--198, 2018. [Online]. Available:
  \url{https://www.osapublishing.org/optica/abstract.cfm?uri=optica-5-2-193}
\BIBentrySTDinterwordspacing

\bibitem{liu2019hybrid}
\BIBentryALTinterwordspacing
J.~Liu, K.~H. Lim, K.~L. Wood, W.~Huang, C.~Guo, and H.-L. Huang, ``Hybrid
  quantum-classical convolutional neural networks,'' \emph{arXiv preprint},
  2019. [Online]. Available: \url{https://arxiv.org/abs/1911.02998}
\BIBentrySTDinterwordspacing

\bibitem{huang2017homomorphic}
\BIBentryALTinterwordspacing
H.-L. Huang, Y.-W. Zhao, T.~Li, F.-G. Li, Y.-T. Du, X.-Q. Fu, S.~Zhang,
  X.~Wang, and W.-S. Bao, ``Homomorphic encryption experiments on ibm's cloud
  quantum computing platform,'' \emph{Frontiers of Physics}, vol.~12, no.~1, p.
  120305, 2017. [Online]. Available:
  \url{https://link.springer.com/article/10.1007/s11467-016-0643-9}
\BIBentrySTDinterwordspacing

\bibitem{huang2020experimental}
H.-L. Huang, Y.~Du, M.~Gong, Y.~Zhao, Y.~Wu, C.~Wang, S.~Li, F.~Liang, J.~Lin,
  Y.~Xu \emph{et~al.}, ``Experimental quantum generative adversarial networks
  for image generation,'' \emph{arXiv:2010.06201}, 2020.

\bibitem{huang2018demonstration2}
H.-L. Huang, A.~K. Goswami, W.-S. Bao, and P.~K. Panigrahi, ``Demonstration of
  essentiality of entanglement in a deutsch-like quantum algorithm,''
  \emph{SCIENCE CHINA Physics, Mechanics \& Astronomy}, vol.~61, no. 060311,
  2018.

\bibitem{huang2021emulating}
H.-L. Huang, M.~Naro{\.z}niak, F.~Liang, Y.~Zhao, A.~D. Castellano, M.~Gong,
  Y.~Wu, S.~Wang, J.~Lin, Y.~Xu \emph{et~al.}, ``Emulating quantum
  teleportation of a majorana zero mode qubit,'' \emph{Physical Review
  Letters}, vol. 126, no.~9, p. 090502, 2021.

\bibitem{simon1997power}
\BIBentryALTinterwordspacing
D.~R. Simon, ``On the power of quantum computation,'' \emph{SIAM J. Comput.},
  vol.~26, no.~5, pp. 1474--1483, July 1997. [Online]. Available:
  \url{https://doi.org/10.1137/S0097539796298637}
\BIBentrySTDinterwordspacing

\bibitem{qRecommendation}
\BIBentryALTinterwordspacing
I.~Kerenidis and A.~Prakash, ``Quantum recommendation systems,'' in \emph{8th
  Innovations Theoretical Computer Sci. Conf.}, ser. Leibniz International
  Proceedings in Informatics (LIPIcs), vol.~67, Berkeley, CA, USA, Jan. 2017,
  pp. 49:1--49:21. [Online]. Available:
  \url{http://drops.dagstuhl.de/opus/volltexte/2017/8154}
\BIBentrySTDinterwordspacing

\bibitem{qi_Recommendation}
\BIBentryALTinterwordspacing
E.~Tang, ``A quantum-inspired classical algorithm for recommendation systems,''
  in \emph{Proc. 51st Annual ACM SIGACT Symposium Theory Computing},
  vol.~25.\hskip 1em plus 0.5em minus 0.4em\relax New York, NY, USA: ACM, June
  2019, pp. 217--228. [Online]. Available:
  \url{https://doi.org/10.1145/3313276.3316310}
\BIBentrySTDinterwordspacing

\bibitem{FKV}
\BIBentryALTinterwordspacing
A.~Frieze, R.~Kannan, and S.~Vempala, ``Fast monte-carlo algorithms for finding
  low-rank approximations,'' \emph{J. Assoc. Comput. Mach.}, vol.~51, no.~6,
  pp. 1025--1041, Nov. 2004. [Online]. Available:
  \url{http://doi.acm.org/10.1145/1039488.1039494}
\BIBentrySTDinterwordspacing

\bibitem{qpca}
\BIBentryALTinterwordspacing
S.~Lloyd, M.~Mohseni, and P.~Rebentrost, ``Quantum principal component
  analysis,'' \emph{Nat. Phys.}, vol.~10, no.~9, p. 631–633, July 2014.
  [Online]. Available: \url{https://doi.org/10.1038/nphys3029}
\BIBentrySTDinterwordspacing

\bibitem{qsc}
\BIBentryALTinterwordspacing
S.~Lloyd, M.~Mohseni, and P.~Rebentrost, ``Quantum algorithms for supervised
  and unsupervised machine learning,'' \emph{arXiv preprint}, Nov. 2013.
  [Online]. Available: \url{https://arxiv.org/abs/1307.0411}
\BIBentrySTDinterwordspacing

\bibitem{qi_PCA}
\BIBentryALTinterwordspacing
E.~Tang, ``Quantum-inspired classical algorithms for principal component
  analysis and supervised clustering,'' \emph{arXiv preprint}, Oct. 2018.
  [Online]. Available: \url{http://arxiv.org/abs/1811.00414}
\BIBentrySTDinterwordspacing

\bibitem{regression}
\BIBentryALTinterwordspacing
A.~Gily{\'{e}}n, S.~Lloyd, and E.~Tang, ``Quantum-inspired low-rank stochastic
  regression with logarithmic dependence on the dimension,'' \emph{arXiv
  preprint}, Nov. 2018. [Online]. Available:
  \url{http://arxiv.org/abs/1811.04909}
\BIBentrySTDinterwordspacing

\bibitem{chia}
\BIBentryALTinterwordspacing
N.-H. Chia, H.-H. Lin, and C.~Wang, ``Quantum-inspired sublinear classical
  algorithms for solving low-rank linear systems,'' \emph{arXiv preprint}, Nov.
  2018. [Online]. Available: \url{https://arxiv.org/abs/1811.04852}
\BIBentrySTDinterwordspacing

\bibitem{hhl}
A.~W. Harrow, A.~Hassidim, and S.~Lloyd, ``Quantum algorithm for linear systems
  of equations,'' \emph{Phys. Rev. Lett.}, vol. 103, no.~15, p. 150502, Oct.
  2009.

\bibitem{qip}
\BIBentryALTinterwordspacing
J.~M. Arrazola, A.~Delgado, B.~R. Bardhan, and S.~Lloyd, ``Quantum-inspired
  algorithms in practice,'' \emph{{Quantum}}, vol.~4, p. 307, Aug. 2020.
  [Online]. Available: \url{https://doi.org/10.22331/q-2020-08-13-307}
\BIBentrySTDinterwordspacing

\bibitem{face}
\BIBentryALTinterwordspacing
P.~J. Phillips, ``Support vector machines applied to face recognition,'' in
  \emph{Advances Neural Inform. Processing Systems}, vol.~48, no. 6241,
  Gaithersburg, MD, USA, Nov. 1999, pp. 803--809. [Online]. Available:
  \url{https://doi.org/10.6028/nist.ir.6241}
\BIBentrySTDinterwordspacing

\bibitem{lssvm}
\BIBentryALTinterwordspacing
J.~A.~K. Suykens and J.~Vandewalle, ``Least squares support vector machine
  classifiers,'' \emph{Neural Process. Lett.}, vol.~9, no.~3, pp. 293--300,
  June 1999. [Online]. Available: \url{https://doi.org/10.1023/A:1018628609742}
\BIBentrySTDinterwordspacing

\bibitem{platt_sequential_1998}
\BIBentryALTinterwordspacing
J.~Platt, ``\BIBforeignlanguage{en-US}{Sequential {Minimal} {Optimization}: {A}
  {Fast} {Algorithm} for {Training} {Support} {Vector} {Machines}},'' Apr.
  1998. [Online]. Available:
  \url{https://www.microsoft.com/en-us/research/publication/sequential-minimal-optimization-a-fast-algorithm-for-training-support-vector-machines/}
\BIBentrySTDinterwordspacing

\bibitem{graf_parallel_2005}
\BIBentryALTinterwordspacing
H.~P. Graf, E.~Cosatto, L.~Bottou, I.~Dourdanovic, and V.~Vapnik, ``Parallel
  {Support} {Vector} {Machines}: {The} {Cascade} {SVM},'' in \emph{Advances in
  {Neural} {Information} {Processing} {Systems} 17}, L.~K. Saul, Y.~Weiss, and
  L.~Bottou, Eds.\hskip 1em plus 0.5em minus 0.4em\relax MIT Press, 2005, pp.
  521--528. [Online]. Available:
  \url{http://papers.nips.cc/paper/2608-parallel-support-vector-machines-the-cascade-svm.pdf}
\BIBentrySTDinterwordspacing

\bibitem{Markov1}
\BIBentryALTinterwordspacing
J.~Xu, Y.~Y. Tang, B.~Zou, Z.~Xu, L.~Li, Y.~Lu, and B.~Zhang, ``The
  generalization ability of svm classification based on markov sampling,''
  \emph{IEEE transactions on cybernetics}, vol.~45, no.~6, pp. 1169--1179,
  2014. [Online]. Available:
  \url{https://ieeexplore.ieee.org/abstract/document/6881630}
\BIBentrySTDinterwordspacing

\bibitem{Markov2}
\BIBentryALTinterwordspacing
B.~Zou, C.~Xu, Y.~Lu, Y.~Y. Tang, J.~Xu, and X.~You, ``$ k $-times markov
  sampling for svmc,'' \emph{IEEE transactions on neural networks and learning
  systems}, vol.~29, no.~4, pp. 1328--1341, 2017. [Online]. Available:
  \url{https://ieeexplore.ieee.org/abstract/document/7993056/}
\BIBentrySTDinterwordspacing

\bibitem{qsvm}
\BIBentryALTinterwordspacing
P.~Rebentrost, M.~Mohseni, and S.~Lloyd, ``Quantum support vector machine for
  big data classification,'' \emph{Phys. Rev. Lett.}, vol. 113, p. 130503,
  Sept. 2014. [Online]. Available:
  \url{https://link.aps.org/doi/10.1103/PhysRevLett.113.130503}
\BIBentrySTDinterwordspacing

\bibitem{bezanson2017julia}
\BIBentryALTinterwordspacing
J.~Bezanson, A.~Edelman, S.~Karpinski, and V.~B. Shah, ``Julia: A fresh
  approach to numerical computing,'' \emph{SIAM review}, vol.~59, no.~1, pp.
  65--98, 2017. [Online]. Available: \url{https://doi.org/10.1137/141000671}
\BIBentrySTDinterwordspacing

\bibitem{CC01a}
C.-C. Chang and C.-J. Lin, ``{LIBSVM}: A library for support vector machines,''
  \emph{ACM Transactions on Intelligent Systems and Technology}, vol.~2, pp.
  27:1--27:27, 2011, software available at
  \url{http://www.csie.ntu.edu.tw/~cjlin/libsvm}.

\bibitem{sampling_tech}
\BIBentryALTinterwordspacing
D.~Achlioptas, F.~McSherry, and B.~Sch{\"o}lkopf, ``Sampling techniques for
  kernel methods,'' in \emph{Advances Neural Inform. Processing Systems}, T.~G.
  Dietterich, S.~Becker, and Z.~Ghahramani, Eds.\hskip 1em plus 0.5em minus
  0.4em\relax Vancouver, British Columbia, Canada: MIT Press, Dec. 2002, pp.
  335--342. [Online]. Available:
  \url{https://papers.nips.cc/paper/2072-sampling-techniques-for-kernel-methods}
\BIBentrySTDinterwordspacing

\bibitem{book2}
\BIBentryALTinterwordspacing
L.~Wang, \emph{Support Vector Machines for Signal Processing}, 1st~ed.\hskip
  1em plus 0.5em minus 0.4em\relax The Netherlands: Springer, Berlin,
  Heidelberg, 2005, ch.~15, pp. 321--342. [Online]. Available:
  \url{https://doi.org/10.1007/b95439}
\BIBentrySTDinterwordspacing

\bibitem{book}
\BIBentryALTinterwordspacing
L.~Wang, \emph{Multiple Model Estimation for Nonlinear Classification},
  1st~ed.\hskip 1em plus 0.5em minus 0.4em\relax The Netherlands: Springer,
  Berlin, Heidelberg, 2005, ch.~2, pp. 49--76. [Online]. Available:
  \url{https://doi.org/10.1007/b95439}
\BIBentrySTDinterwordspacing

\bibitem{frugal_2018}
\BIBentryALTinterwordspacing
I.~L. Markov, A.~Fatima, S.~V. Isakov, and S.~Boixo,
  ``\BIBforeignlanguage{en}{Quantum {Supremacy} {Is} {Both} {Closer} and
  {Farther} than {It} {Appears}},'' \emph{\BIBforeignlanguage{en}{arXiv
  preprint}}, Sep. 2018. [Online]. Available:
  \url{http://arxiv.org/abs/1807.10749}
\BIBentrySTDinterwordspacing

\end{thebibliography}

\IEEEoverridecommandlockouts
\begin{IEEEbiography}[{\includegraphics[width=1in,height=1.25in,clip,keepaspectratio]{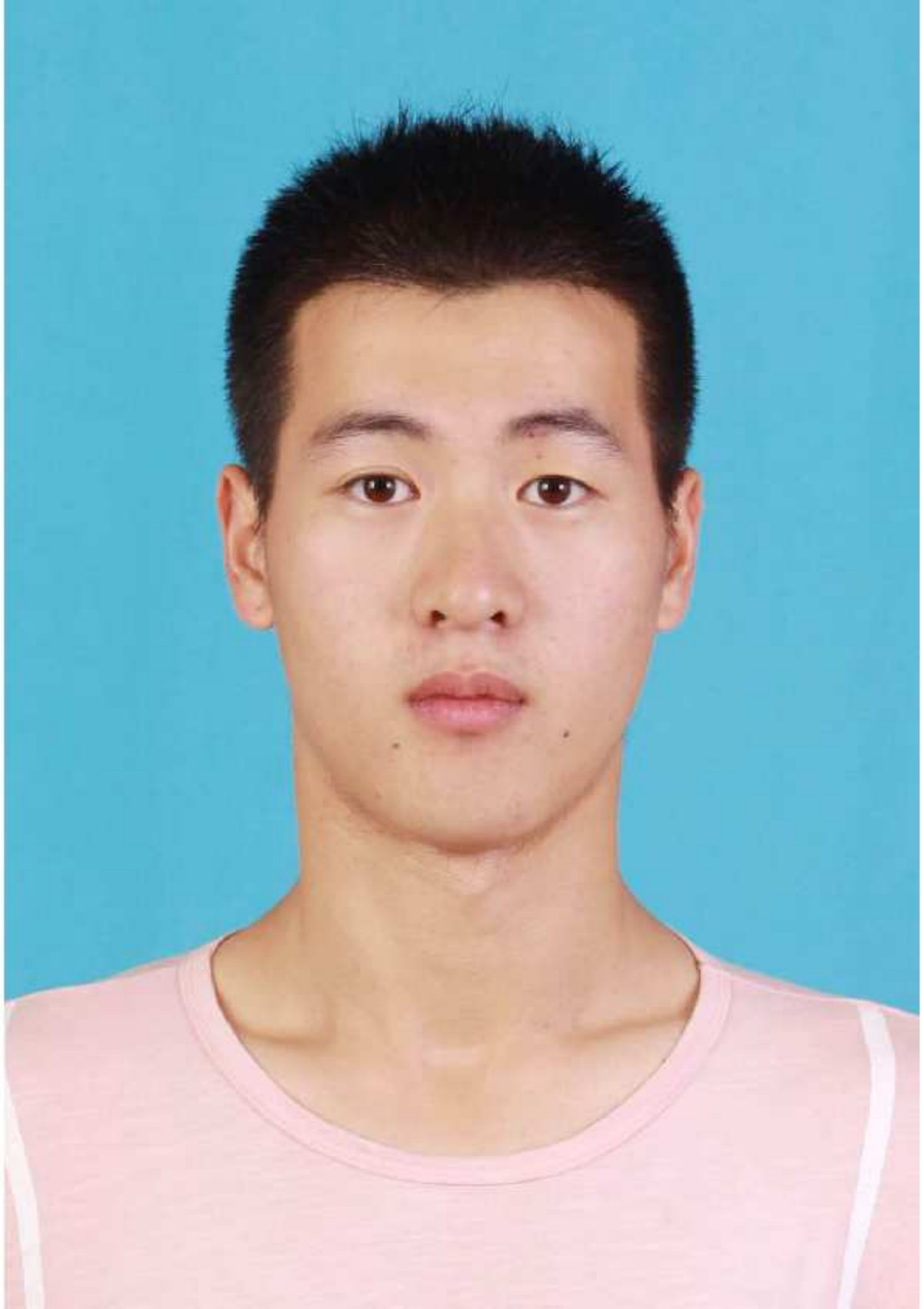}}]{Chen Ding}
received the B.S. degree from University of Science and Technology of China, Hefei, China, in 2019.

He is currently a graduate student in CAS Centre for Excellence and Synergetic Innovation Centre in Quantum Information and Quantum Physics.
His current research interests include quantum machine learning, quantum-inspired algorithm designing and variational quantum computing.
\end{IEEEbiography}

\begin{IEEEbiography}[{\includegraphics[width=1in,height=1.25in,clip,keepaspectratio]{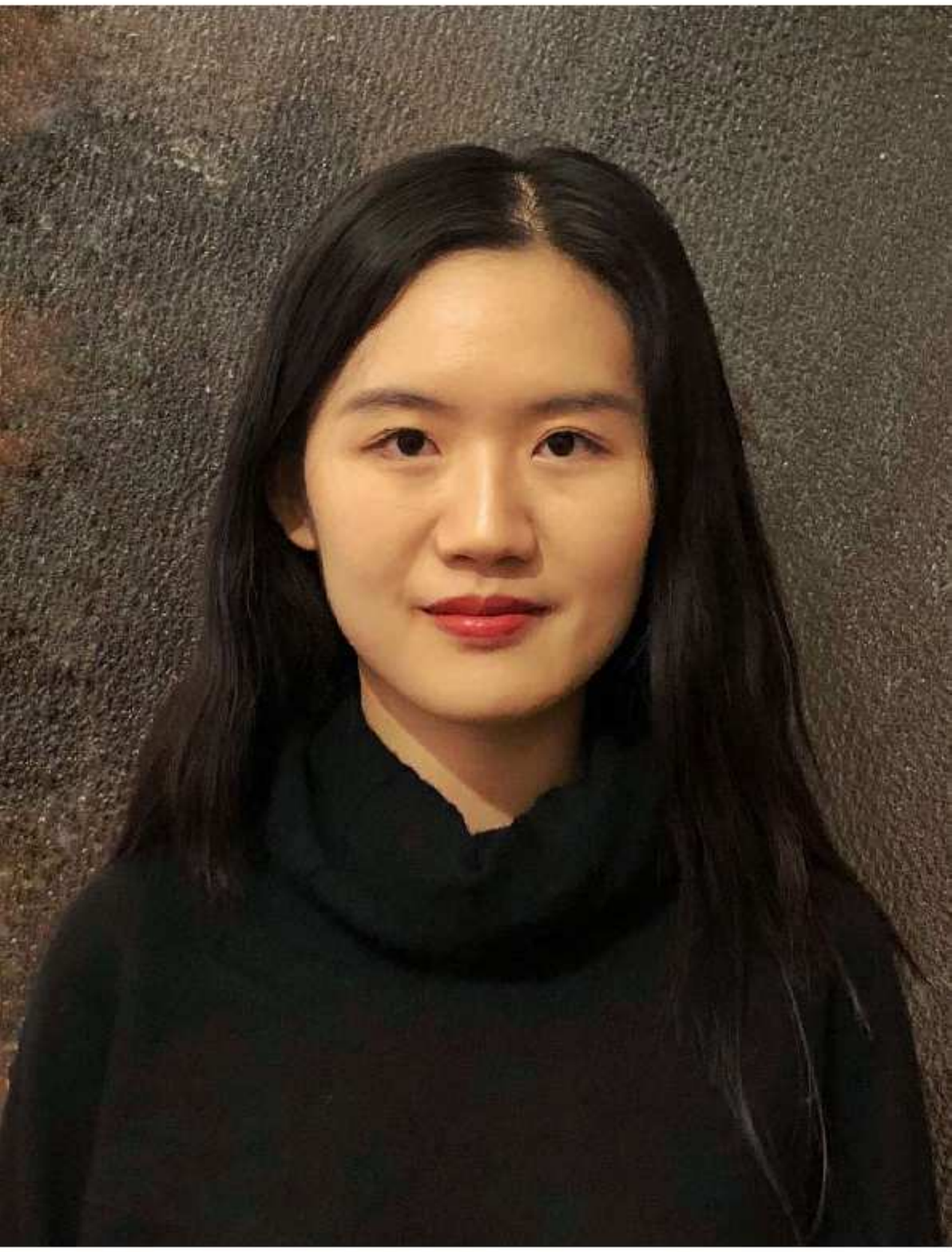}}]{Tian-Yi Bao} received the B.S. degree from University of Michigan, Ann Arbor, USA, in 2020.

She is currently a graduate student in Oxford University. Her current research interests include the machine learning and human-computer interaction.
\end{IEEEbiography}

\begin{IEEEbiography}[{\includegraphics[width=1in,height=1.25in,clip,keepaspectratio]{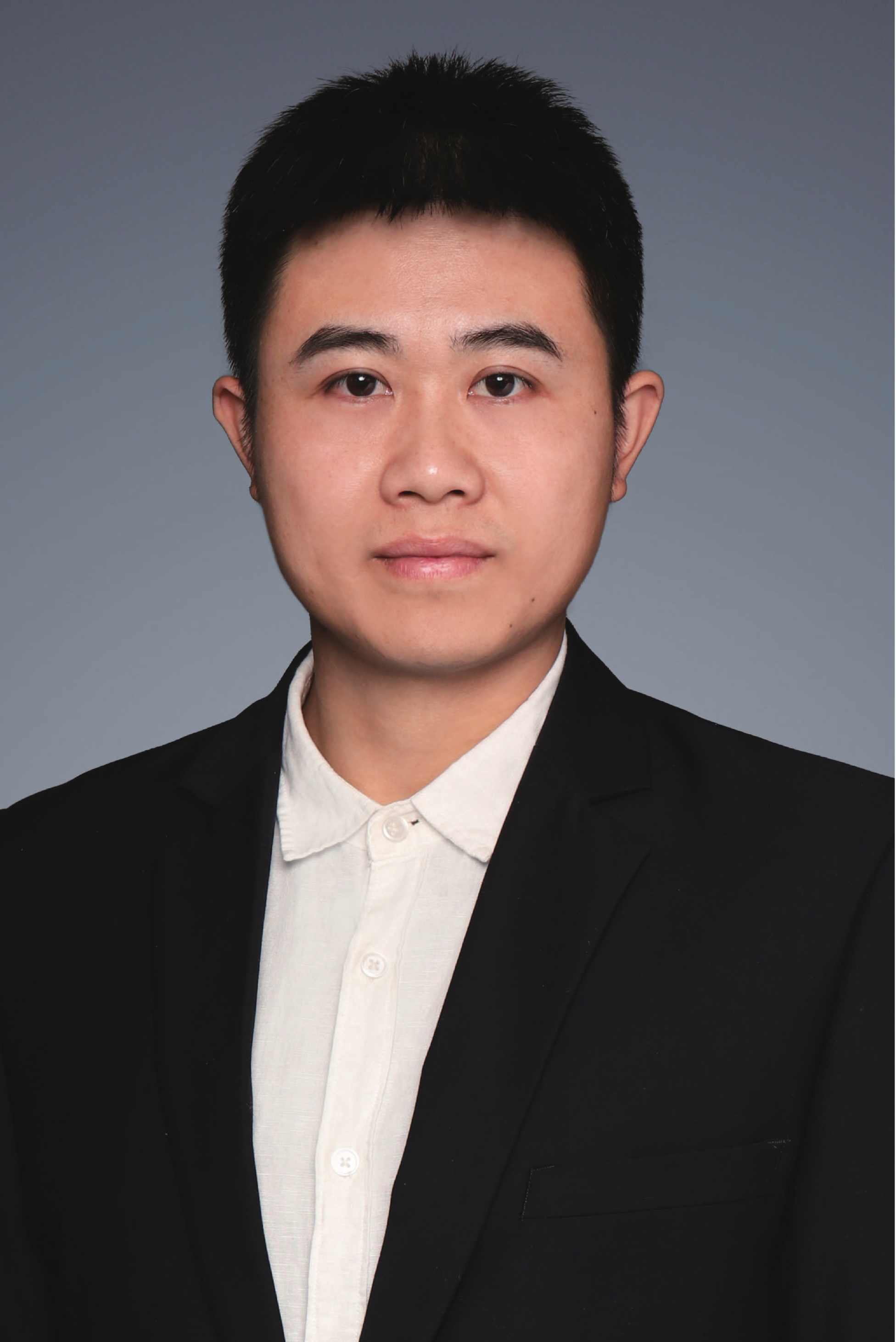}}]{He-Liang Huang}
received the Ph.D. degree from the University of Science and Technology of China, Hefei, China, in 2018.

He is currently an Assistant Professor of Henan Key Laboratory of Quantum Information and Cryptography, Zhengzhou, China, and the Postdoctoral Fellow of University of Science and Technology of China, Hefei, China. He has authored or co-authored over 30 papers in refereed international journals and co-authored 1 book. His current research interests include secure cloud quantum computing, big data quantum computing, and the physical implementation of quantum computing architectures, in particular using linear optical and superconducting systems.
\end{IEEEbiography}

\end{document}